\documentclass{article}

\usepackage[accepted]{icml2022}

\usepackage[utf8]{inputenc} \usepackage[T1]{fontenc}    \usepackage{hyperref}       \usepackage{url}            \usepackage{booktabs}       \usepackage{nicefrac}       \usepackage{microtype}      \usepackage{xcolor}         \usepackage{amsmath,amsthm,amssymb,amsfonts,mathtools,bm,stmaryrd} \usepackage{multirow}
\usepackage{pifont}
\usepackage{bbding}
\usepackage{placeins}
\usepackage{fontawesome}
\usepackage{caption}
\usepackage{pgfplots}
\pgfplotsset{compat=1.16}
\usepgfplotslibrary{fillbetween}

\usepackage{orcidlink}

\newcommand{\Input}{\item[\textbf{Input:}]}
\newcommand{\Output}{\item[\textbf{Output:}]}
\newcommand{\Setup}{\item[\textbf{Setup:}]}
\newcommand{\State}{\STATE}
\newcommand{\Comment}{\COMMENT}
\newcommand{\Return}{\textbf{return }}
\newcommand{\For}{\FOR}
\newcommand{\EndFor}{\ENDFOR}

\newcommand{\tcaption}[1]{\caption{#1} \vskip.3em}

\newcommand{\defeq}{\mathrel{\mathop:}=}

\newcommand{\rl}{\mathsf{ReLU}}

\newcommand{\round}[1]{\lfloor #1 \rceil}
\newcommand{\floor}[1]{\lfloor #1 \rfloor}
\newcommand{\ceil}[1]{\lceil #1 \rceil}

\newcommand{\fpshare}[1]{\ensuremath{\langle #1; f \rangle^{\mathsf{A}}}}
\newcommand{\intshare}[1]{\ensuremath{\langle #1 \rangle^{\mathsf{A}}}}
\newcommand{\binshare}[1]{\ensuremath{\langle #1 \rangle^{\mathsf{B}}}}

\newcommand{\share}[1]{\intshare{#1}}

\newcommand{\asn}{\leftarrow}

\DeclarePairedDelimiter{\iverson}{\llbracket}{\rrbracket}

\newcommand{\tock}{{\ding{56}}}
\newcommand{\tick}{{\ding{52}}}

\newcommand{\yes}{{\textcolor{red!75!}{\faBatteryQuarter}}}
\newcommand{\no}{{\textcolor{-red!75!green!50!blue}{\faBatteryThreeQuarters}}}
\newcommand{\soso}{{\textcolor{orange!50!white}{\faBatteryHalf}}}
\newcommand{\local}{{\textcolor{-red!75!green!50!blue}{\faBatteryFull}}}

\DeclareMathOperator*{\argmin}{arg\,min}

\renewcommand{\Re}{\mathbb{R}}    \newcommand{\Ze}{\mathbb{Z}}      

\newcommand{\uniform}{\mathcal{U}}
\newcommand{\bernoulli}{\mathcal{B}}

\newcommand{\T}{\top}

\definecolor{colour1}{HTML}{FF1F5B}
\definecolor{colour2}{HTML}{00CD6C}
\definecolor{colour3}{HTML}{009ADE}
\definecolor{colour4}{HTML}{AF58BA}
\definecolor{colour5}{HTML}{FFC61E}
\definecolor{colour6}{HTML}{F28522}
\definecolor{colour7}{HTML}{A0B1BA}
\definecolor{colour8}{HTML}{A6761D}

\newcommand{\va}{\mathbf{a}}
\newcommand{\vb}{\mathbf{b}}
\newcommand{\vc}{\mathbf{c}}
\newcommand{\Enc}{\mathsf{Enc}}
\newcommand{\Dec}{\mathsf{Dec}}

\theoremstyle{plain}
\newtheorem{theorem}{Theorem}[section]
\newtheorem{proposition}[theorem]{Proposition}
\newtheorem{lemma}[theorem]{Lemma}

\theoremstyle{definition}

\theoremstyle{remark}

\newcommand{\appref}[1]{Appendix~\ref{#1}}

\newcommand{\switchref}[1]{\appref{#1}}

\newcommand{\PRG}{\mathsf{PRG}}

\title{Secure Quantized Training for Deep Learning}

\hypersetup{colorlinks=false} \author{
  Marcel Keller$^\dagger$~\orcidlink{0000-0003-2261-9376}\hspace{5em}Ke Sun$^\dagger$$^\star$~\orcidlink{0000-0001-6263-7355}\\
  $^\dagger$CSIRO's Data61\hspace{2em}$^\star$The Australian National University\\
  \texttt{marcel.keller@data61.csiro.au}\hspace{2em}\texttt{sunk@ieee.org}
}
\hypersetup{colorlinks=true, citecolor=blue}

\makeatletter
\begin{document}

\twocolumn[
\icmltitle{\@title}

\begin{icmlauthorlist}
\icmlauthor{Marcel Keller}{d61}
\icmlauthor{Ke Sun}{d61,anu}
\end{icmlauthorlist}

\icmlaffiliation{d61}{CSIRO's Data61, Sydney, Australia}
\icmlaffiliation{anu}{The Australian National University}

\icmlcorrespondingauthor{Marcel Keller}{marcel.keller@data61.csiro.au}
\icmlcorrespondingauthor{Ke Sun}{ke.sun@data61.csiro.au}

\icmlkeywords{Privacy-preserving machine learning, secure multi-party
  computation}

\vskip 0.3in
]

\printAffiliationsAndNotice{}

\begin{abstract}
  We implement training of neural networks in secure
  multi-party computation (MPC) using quantization commonly used in said
  setting. We are the first to present
  an MNIST classifier purely trained in MPC that comes within 0.2 percent
  of the accuracy of the same convolutional neural network trained via
  plaintext computation. More
  concretely, we have trained a network with two convolutional and two
  dense layers to 99.2\% accuracy in 3.5 hours (under one hour for 99\% accuracy).
  We have also implemented AlexNet for CIFAR-10, which converges in a few hours.
  We develop novel protocols for exponentiation and inverse square root.
  Finally, we present experiments in a
  range of MPC security models for up to ten parties, both with honest
  and dishonest majority as well as semi-honest and malicious security.
\end{abstract}

\section{Introduction}

Secure multi-party computation (MPC) is a cryptographic technique that
allows a set of parties to compute a public output on private inputs
without revealing the inputs or any intermediate results. This makes
it a potential solution to federated learning where the sample data
stays private and only the model or even only inference results are
revealed.

Imagine a set of healthcare providers holding sensitive patient
data. MPC allows them to collaboratively train a model. This model
could then either be released or even kept private for inference using
MPC again. See Figure~\ref{fig:concept} for an illustration.
Note that MPC is oblivious to how the input data is split among
participants, that is, it can be used for the horizontal as well as
the vertical case.

A more conceptual example is the well-known \emph{millionaires'
problem} where two people want to find out who is richer without
revealing their wealth. There is clearly a difference between the one
bit of information desired and the full figures.

There has been a sustained interest in applying secure computation to
machine learning and neural networks going back to at least
\citet{barni2006privacy}. More recent advances in practical MPC have
led to an increased effort in implementing both inference and training.

\usetikzlibrary{calc}
\usetikzlibrary{graphs}

\begin{figure}[b]
  \centering
  \newlength{\server}
  \newlength{\hosp}
  \setlength{\server}{0.5cm}
  \setlength{\hosp}{1.5cm}
  \begin{tikzpicture}[x=.2cm,y=.2cm]
    \coordinate (A) at (0,0);
    \coordinate (B) at (5,4);
    \node (angie) at (A) {\includegraphics[width=\server]{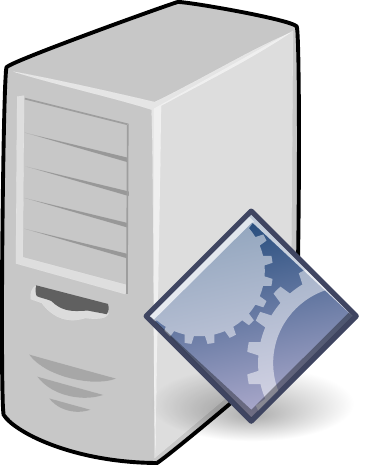}};
    \node (keith) at (B) {\includegraphics[width=\server]{server}};
    \node (ed) at ($ (A) ! .5 ! (B) ! {sin(60)*2} ! 90:(B) $)
    {\includegraphics[width=\server]{server}};
    \graph{(angie) <-> (keith) <-> (ed) <-> (angie)};
    \coordinate (C) at (-25, 0);
    \coordinate (D) at (-19, 6);
    \node (client1) at (C) {\includegraphics[width=\hosp]{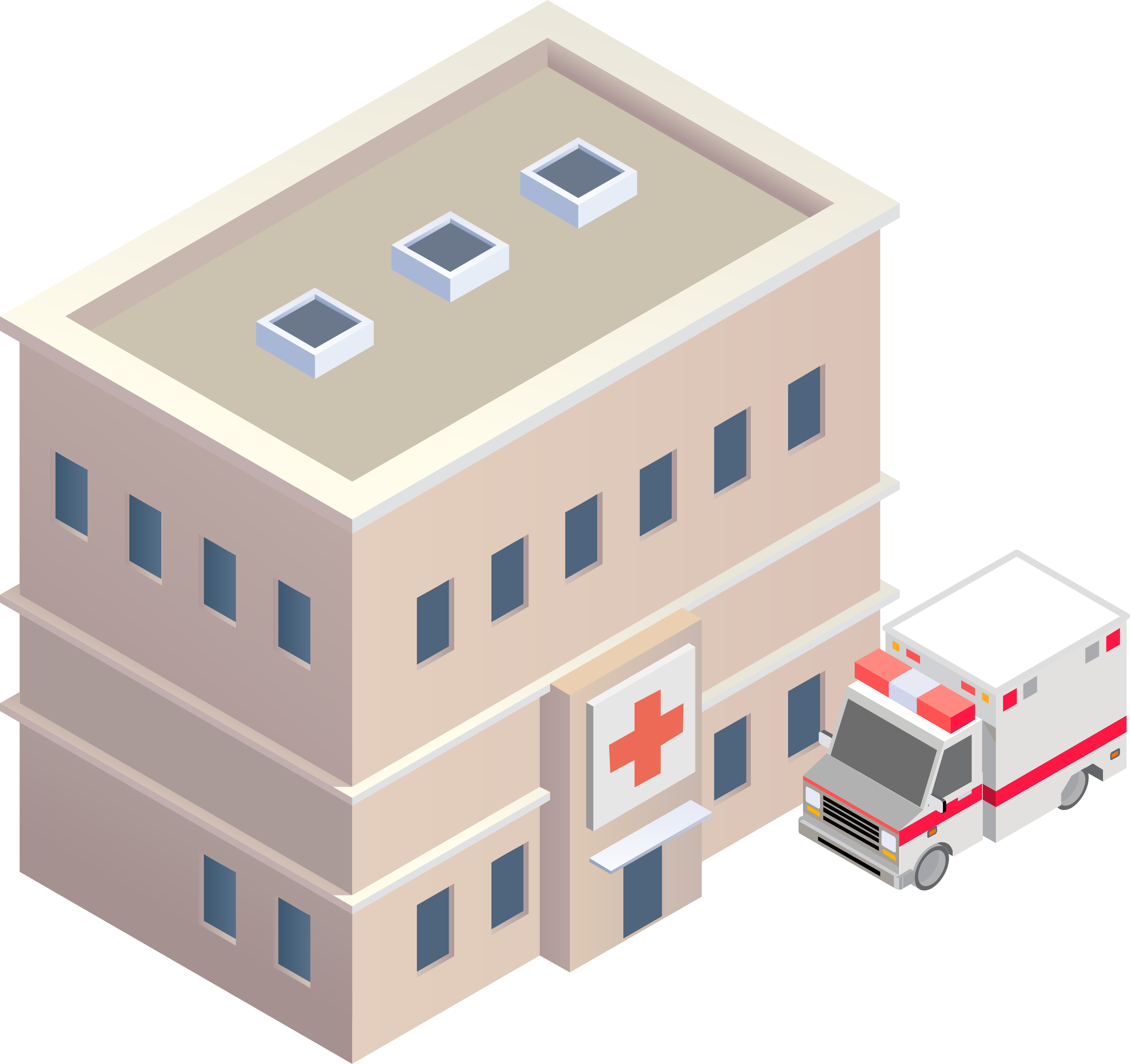}};
    \node (client2) at (D) {\includegraphics[width=\hosp]{hospital}};
    \graph{(ed) <-> (client1) <-> (angie)};
    \graph{(keith) <-> (client1)};
    \graph{(ed) <-> (client2) <-> (angie)};
    \graph{(keith) <-> (client2)};
  \end{tikzpicture}
  \caption{Outsourced computation: Data holders (on the left)
    secret-share their data to a number of computing parties (on the
    right), who then return the desired result (e.g., a model or
    inference results on further queries). All communication except
    the outputs is secret-shared and thus secure if no two computing
    parties collude.}
  \label{fig:concept}
\end{figure}

A number of works such as \citet{SP:MohZha17}, \citet{CCS:MohRin18},
\citet{PoPETS:WagGupCha19}, \citet{PoPETS:WTBKMR21} implement neural
network training with MPC at
least in parts. However, for the task of classifying MNIST digits,
they either give accuracy figures below 95\%
or figures obtained using plaintext
training. For the latter case, the works do not clarify how close the
computation for plaintext training matches the lower precision and
other differences in the MPC setting.
\citet{CCS:ASKG19} achieved an accuracy of 99.38\% in a comparable setting for a
convolutional neural network with more channels than we use. However,
their actual implementation only uses dense layers,
and we achieve comparable accuracy in this model.
All works use quantization in the sense that a fractional number
$x$ is represented as $\round{x \cdot 2^{f}}$ where $\round{\cdot}$
denotes rounding to the nearest integer, and $f\in\Ze$ ($\Ze$: the set of integers)
is a precision parameter. This makes addition
considerably faster in the secure computation setting because it
reduces to integer addition (at the cost of compounding rounding
errors). Furthermore, some of the works
suggest replacing the softmax function that uses exponentiation with a
ReLU-based replacement.
\citet{keller2020effectiveness} discovered that this softmax
replacement deteriorates accuracy in multilayer perceptrons to the
extent that it does not justify the efficiency gains.

The concurrent work of \citet{tan2021cryptgpu}
gives some figures on the learning curves when trained with secure
computation. However, they stop at five epochs for MNIST training
and achieve 94\% accuracy, whereas we present the figures up to
50 epochs and 99.2\% accuracy using AMSGrad.
Our implementation using stochastic gradient descent (SGD) is 40\% faster than
theirs. Note that we use the CPU of one AWS {c5.9xlarge}
instance per party whereas \citeauthor{tan2021cryptgpu} use one NVIDIA Tesla V100 GPU
per party. We believe this somewhat counter-intuitive result comes
from MPC heavily relying on
communication, which is an aspect where GPUs do not have an advantage
over CPUs. While GPUs are very efficient in computing on data
in their cache, it is no more efficient to transfer data in and out of
these caches.

In this paper, we present an extensible framework for
secure deep learning based on MP-SPDZ by \citet{CCS:Keller20},
a software library for multi-party computation.\footnote{Code available at \url{https://github.com/data61/MP-SPDZ}.}
Similar to TensorFlow and
PyTorch, our approach allows representing deep learning models as
a succession of layers. We then use this implementation to obtain accuracy
figures for MNIST and CIFAR-10 training by utilizing the MP-SPDZ emulator, which
allows to run the plaintext equivalent of secure computation, that is,
the same algorithms with the same precision. Finally, we run one of
the most
promising instantiations in real secure computation in order to benchmark
it, confirming the results from the plaintext emulator.
We use less sophisticated datasets because the relatively high cost of MPC
makes is less feasible to use datasets with millions of examples
rather than tens of thousands.

For our implementation, we have designed novel protocols for
exponentiation and inverse square root computation. Both play an
important role in deep learning, the first in softmax, and the second
in the Adam and AMSGrad optimizers. Our protocols are based on
mixed-circuit computation, that is, we combine a protocol that works
over $\mathbb{Z}_M$ for some large modulus $M$ and thus integer-like
computation with a protocol in the same security model that works over
$\mathbb{Z}_2$, which allows computing binary circuits.\footnote{$\Ze_M = \Ze/M\Ze$ denotes integer computation
  modulo $M$.}

There are several projects that integrate secure computation
directly into popular machine learning frameworks such as CrypTen by
\citet{crypten}, PySyft by \citet{pysyft}, and TF Encrypted by
\citet{tf-encrypted}. Our approach differs from all of them by running
the protocol as native CPU code (implemented using \texttt{C++}). This allows for
much faster execution. For example, CrypTen provides an MNIST training
example (\verb+mpc_autograd_cnn+), which we adapted to full LeNet~\citep{lecun1998gradient}
training. It took over three hours to run one epoch on one
machine. In comparison, our
implementation takes 11 minutes to run one epoch with the full dataset
of 60,000 samples with the same hardware across instances.

An alternative to MPC would be using purely homomorphic
encryption. \citet{glyph} trained an MNIST network to 98.6\% accuracy in 8
days whereas we achieved the same accuracy in less than 5 days when using a
two-party protocol based on homomorphic encryption.

A number of works \citep{USENIX:JuvVaiCha18,USENIX:MLSZP20,CCS:RRKCGRS20} only
implemented deep learning inference. None of them implements all building
blocks necessary for training. It is therefore not possible to make any
statement on the training performance of these works.

Another line of work (e.g., \citealt{secure-tf}) uses trusted execution
environments that provide computation outside the reach of the
operating system. This is a different security model to multi-party
computation that works with distributing the information among several
entities.

The rest of this paper is structured as follows: We highlight some
secure computation aspects in Section \ref{sec:main-blocks} and do the
same for machine learning in Section \ref{sec:ml}.  Section
\ref{sec:error} analyzes the error introduced by the quantization.
Finally, we present our implementation and our experimental results
for MNIST and CIFAR-10 classification in Section
\ref{sec:implementation}.

 \section{Secure Computation Building Blocks}
\label{sec:main-blocks}

Our secure implementation has two layers of abstraction. The lower
level is concerned with a few basic operations that depend on the
actual protocol and security model. The basic operations include
integer addition and multiplication, input, output, and domain
conversion, that is converting between integer-focused and bit-focused
computation.  Appendices~\ref{sec:mpc} and \ref{sec:he} give
the details of our honest-majority three-party protocol and our
dishonest-majority protocol, respectively.
Honest and dishonest majority refer to how many computing parties are
being trusted. For an honest majority, strictly more than half parties
must follow the protocol and not reveal information to anyone
else. This setting allows for more efficient protocols while the
dishonest-majority setting requires more expensive cryptographic
schemes such as homomorphic encryption to be used. See
\citet{CCS:Keller20} for a cost comparison between a range of protocols.
The higher level of abstraction is common
to all protocol as it builds in a generic way on the primitives in the
lower level. Notable operations on this level are division,
exponentiation, and inverse square root. Appendix \ref{sec:blocks}
provides more details.

In this section, we focus on two aspects: How to implement
fractional-number computation with integers, and how to implement
exponentiation. The former plays a crucial role in
efficient secure computation, and the latter presents a
novel protocol.
We also contribute a novel protocol for inverse square root
computation, which is deferred to Appendix \ref{sec:blocks} due to space constraints.

\subsection{Quantization}

While \citet{NDSS:ABZS13} showed that it is possible to implement
floating-point computation, the cost is far higher than integer
computation. It is therefore common to represent fractional numbers
using quantization (also called fixed-point representation) as
suggested by \citet{FC:CatSax10}. A real number $x$ is represented as
$Q^f(x)\defeq\round{x \cdot 2^{f}}$ where $f$ is a positive integer specifying the
precision. The linearity of the representation allows to compute
addition by simply adding the representing integers. Multiplication
however requires adjusting the result because it will have twice the
precision: $(x \cdot 2^f) \cdot (y \cdot 2^f) = xy \cdot 2^{2f}$. There
are two ways to rectify this:
\begin{itemize}
\item An obvious correction would be to shift the result by $f$ bits
  after adding $2^{f-1}$ to the integer representation,
as $\round{xy\cdot{}2^{f}}=\floor{2^{-f}\cdot (xy\cdot2^{2f} + 2^{f-1})}$,
where $\floor{\cdot}$ is the floor function. This ensures
  rounding to the nearest number possible in the representation, with
  the tie being broken by rounding up.
\item However, \citeauthor{FC:CatSax10} found that in the context of
  secure computation it is more efficient to use probabilistic
  truncation. This method rounds up or down probabilistically
  depending on the input. For example, probabilistically rounding 0.75
  to an integer would see it rounded up with probability 0.75 and
  down with probability 0.25.
\end{itemize}
Our quantization scheme is
related to quantized neural networks  (see e.g. \citealt{bnn}).
However, our consideration is not to compress the model, but to improve the computational
speed and to reduce communication.

 \subsection{Exponentiation}
\label{sec:exp}

\begin{figure}[t!]
\begin{minipage}{\columnwidth}
\begin{algorithm}[H]
\caption{Exponentiation with base two \citep{ACNS:AlySma19}}
\label{alg:simple-exp}
\begin{algorithmic}[1]
    \Input{Secret share $\fpshare{x}$ with precision $f$ where
      $-(k - f - 1) < x < k - f - 1$}
    \Output{$\fpshare{2^x}$ with precision $f$ and total bit length $k$}

    \State $\share{s} \asn \fpshare{x}$ < 0
    \Comment{sign}
    \State $\fpshare{x} \asn \fpshare{x} - 2 \cdot \share{s} \cdot
    \fpshare{x}$
    \Comment{absolute value}
    \State $\share{i} \asn \mathsf{Trunc}(\fpshare{x})$
    \Comment{integer component}
    \State $\fpshare{r} \asn \fpshare{x} - \share{i}$
    \Comment{fractional component}
    \State $\ell \asn \ceil{\log_2(k - f)}$
    \State $\share{i_0}, \dots, \share{i_{\ell-1}} \asn
    \mathsf{BitDec}(\share{r}, \ell)$
    \Comment{bit decomposition}
    \State $\share{d} \asn \prod_{j=0}^{\ell-1}(\share{i_j} \cdot 2^{2^{j}}
    + 1 - \share{i_j})$
    \Comment{integer exponentiation}
    \State $\fpshare{u} \asn \mathsf{Approx}_{2^*}(\fpshare{r})$
    \Comment{polynomial approximation}
    \State $\fpshare{g} \asn \fpshare{u} \cdot \share{d}$
    \State \Return $\share{s} \cdot \left(\frac{1}{\fpshare{g}} -
      \fpshare{g} \right) + \fpshare{g}$
    \Comment{correct for sign}
  \end{algorithmic}
\end{algorithm}

 \end{minipage}
\begin{minipage}{\columnwidth}
\begin{algorithm}[H]
\caption{Exponentiation with base two (ours)}
\label{alg:our-exp}
\begin{algorithmic}[1]
  \Input{Secret share $\fpshare{x}$ with precision $f$ where
    $x < k - f - 1$}
  \Output{$\fpshare{2^x}$ with precision $f$ and total bit length $k$}

  \State $\binshare{x_0}, \cdots, \binshare{x_{k-1}} \asn
  \mathsf{A2B}(\fpshare{x})$
  \Comment{bit decomposition}
  \State $\binshare{z} \asn \sum_{i=0}^{k-1}  2^{i-f} \binshare{x_{i}} < -(k-f-1)$
  \Comment{binary comparison with fixed-representation}
  \State $\ell \asn \ceil{\log_2(k - f)}$
  \For{$j = f, \dots, f+\ell-1$}
  \State $\share{x_j} \asn \mathsf{Bit2A}(\binshare{x_j})$
  \EndFor
  \State $\share{d} \asn \prod_{j=0}^{\ell-1}(\share{x_{f + j}} \cdot 2^{2^{j}}
  + 1 - \share{x_{f+j}})$
  \Comment{integer exponentiation}
  \State $\share{r} \asn \mathsf{B2A}(\binshare{x_0}, \dots,
  \binshare{x_{f-1}})$
  \Comment{fractional component}
  \State $\fpshare{u} \asn \mathsf{Approx}_{2^*}(\fpshare{r})$
  \Comment{polynomial approximation}
  \State $\fpshare{g} \asn \fpshare{u} \cdot \share{d}$
  \label{step:g}
  \State $\fpshare{g'} \asn \mathsf{Round}(\fpshare{g}, f + 2^\ell,
  2^\ell)$
  \label{step:gprime}
  \Comment{result for negative input}
  \label{step:trunc}
  \State $\fpshare{h} \asn \mathsf{Bit2A}(\binshare{x_{k-1}}) \cdot
  \left(\fpshare{g'} - \fpshare{g} \right) + \fpshare{g}$
  \Comment{correct for sign}
  \State \Return $(1 - \mathsf{Bit2A}(\binshare{z})) \cdot
  \fpshare{h}$
  \Comment{output 0 if input is too small}
\end{algorithmic}
\end{algorithm}
\end{minipage}
\end{figure}

\begin{algorithm}
  \caption{Procedures in Algorithms \ref{alg:simple-exp} and \ref{alg:our-exp}}
\begin{description}
\item[$\binshare{x_0}, \dots, \binshare{x_{k-1}} \asn
  \mathsf{A2B}(\fpshare{x})$] Arithmetic to binary conversion of the
  $k$ least significant bits, which can be implemented using edaBits
  \citep{C:EGKRS20} or local share conversion
  \citep{NDSS:DemSchZoh15,CCS:ABFKLO18} if available in the underlying
  protocol.
\item[$\intshare{x} \asn \mathsf{Bit2A}(\binshare{x})$] Bit to arithmetic conversion, which can be
  implemented using daBits \citep{INDOCRYPT:RotWoo19}.
\item[$\intshare{x} \asn \mathsf{B2A}(\binshare{x_0}, \dots,
  \binshare{x_k})$] Binary to arithmetic conversion of $k$-bit values,
  which can be implemented using edaBits.
\item[$\fpshare{y} \asn \mathsf{Round}(\fpshare{x}, k, m)$] Truncate a
  $k$-bit value by $m$ bits with either nearest or probabilistic
  rounding, corresponding to a division by $2^m$. Both also benefit
  from mixed-circuit computation as shown by
  \citet{USENIX:DalEscKel21}.
\item[$\fpshare{y} \asn \mathsf{Approx_{2^*}}(\fpshare{x})$]
  Approximate $2^x$ for $x \in [0,1]$ using a Taylor series.
\end{description}
\end{algorithm}

Exponentiation is a core computational module in deep learning
and is useful to evaluate non-linear activation functions such
as the softmax. \citet{SP:MohZha17}
introduced a simplification of softmax to avoid computing
exponentiation. In this section, we present an optimized
exponentiation protocol instead.

As noted
by \citet{ACNS:AlySma19}, exponentiation can by reduced to
exponentiation with base two using $a^x = 2^{x \log_2
  a}$. Furthermore, if the base is public, this reduction only costs
one public-private multiplication, and it introduces an acceptable
error as
\[ 2^{x \log_2 a \pm 2^{-f}} = 2^{x \log_2 a} \cdot 2^{\pm 2^{-f}}. \]
The second multiplicand on the right-hand side is easily seen to be
within $[1 - 2^{-f}, 1 +  2^{-f}]$.

Algorithm \ref{alg:simple-exp} shows the exponentiation method by \citeauthor{ACNS:AlySma19}.
It uses $\share{\cdot}$ to indicate
integer sharing and $\fpshare{\cdot}$ to indicate fixed-point
sharing.
We use the superscript $\mathsf{A}$ to indicate that both secret
sharing are in the arithmetic domain, that is, in $Z_M$ for some large
$M$. This distinguishes them from the bit-wise secret sharing
indicated by $\mathsf{B}$ used below.
The algorithm uses a number of standard components such as
addition, multiplication, comparison, and bit
decomposition. Furthermore, it uses a polynomial approximation of
$2^x$ for $x \in [0,1]$. They propose to use the polynomial $P_{1045}$
by \citet{hart1978computer}. However, we found that Taylor
approximation at 0 leads to better results than $P_{1045}$.

Algorithm \ref{alg:simple-exp} suffers from two
shortcomings. First, it is unnecessarily restrictive on negative
inputs because it uses the absolute value of the input, the
exponentiation of which falls outside the representation range for
values as small as $k - f - 1$ where $k$ is a parameter defining the
range of inputs ($[-2^{k-f-1},2^{k-f-1}]$). Second, the truncation used to find
the integer component resembles bit decomposition in many
protocols. It is therefore wasteful to separate the truncation and the
bit decomposition. Algorithm \ref{alg:our-exp} fixes these two issues.
It also explicitly uses mixed circuits, with binary secrets denoted using $\binshare{\cdot}$.

The general structure of Algorithm \ref{alg:our-exp} is the same as
Algorithm \ref{alg:simple-exp}. It splits the input into an integer
and a fractional part and then computes the exponential separately,
using an exact algorithm for the integer part and an approximation for
the fractional part. It differs in that it uses mixed circuits and it
allows any negative input. The biggest difference however is in step
\ref{step:trunc}, where the exponentiation of a negative number is
not computed by inverting the exponentiation of the absolute
value. Instead, we use the following observation. If $-2^\ell \le x <
0$, $x = -2^\ell + y + r$, where $y = x + 2^\ell - r$ and $r$ is the
fractional component of $x$.
Then,
\[ 2^x = 2^{-2^\ell} \cdot 2^y \cdot 2^r. \]
$2^y$ equals $d$ in Algorithm \ref{alg:our-exp} because $y$
is the composition of the bits used in integer exponentiation,
namely $y = \sum_{i=0}^{\ell-1} x_{f+i}$.
Therefore, $g \approx 2^y \cdot 2^r$, and
\[ g' \approx \frac{g}{2^{2^\ell}} \approx 2^x \]
for $g$ and $g'$ in steps \ref{step:g} and \ref{step:gprime}
because rounding by $2^\ell$ bits implies division by $2^{2^\ell}$.
This saves a relatively expensive division compared to
Algorithm~\ref{alg:simple-exp}.

Table \ref{table:exp} shows how the two algorithms compare in terms of
communication across a number of security models. All figures are
obtained using MP-SPDZ \citep{CCS:Keller20} with $f=16$, $k=31$,
computation modulo $2^{64}$ for three-party computation, and computation
modulo a 128-bit prime for two-party computation (as in Section~\ref{sec:sec-models}).
We use edaBits for mixed-circuit
computation in Algorithm~\ref{alg:our-exp}.  Across all security
models, our algorithm saves about 30 percent in communication despite
the improved input range.
In the context of the figures in Table~\ref{table:sec-models} the improvement
is less than one percent because exponentiation is only used once for every
example per epoch, namely for the computation of softmax.

\begin{table}
  \centering
  \tcaption{Total communication in kbit for Algorithms
    \ref{alg:simple-exp} and \ref{alg:our-exp} across a range of
    security models with one corrupted party. ``SH'' stands for
    semi-honest security and ``Mal.'' for malicious security.}
  \label{table:exp}
  \begin{tabular}{lrrrr}
    \toprule
    & \multicolumn{2}{c}{3 Parties} & \multicolumn{2}{c}{2 Parties} \\
    & SH & Mal. & SH & Mal. \\
    \midrule
    \citet{ACNS:AlySma19}  & 27 & 498 & 1,338 & 214,476\\
    Algorithm \ref{alg:our-exp} (ours)  & 16 & 323 & 813 & 121,747\\
    \bottomrule
  \end{tabular}
\end{table}

 \section{Deep Learning Building Blocks}\label{sec:ml}

\begin{table}[thb]
\centering
\tcaption{Basic operations and their cost.
    The column ``MPC Cost'' shows the roughly approximated computational
    cost in MPC, as compared to integer multiplication. The last
    column displays the computational and communication resource consumption.
    ``\local'' means the cheapest computation; ``\yes'' means the most expensive.}\label{table:basicops}
\begin{tabular}{rccc}
    \toprule
    Operation & MPC Cost ($\approx$) & Efficiency \\
    \hline
    $\iverson{x>y}$ & $10^1$ & \soso \\
    $x\pm{y}$ & $10^{-1}$ & \local \\
    $x\times{y}$ (integer) & $10^0$ & \no \\
    $x\times{y}$ (fractional) & $10^1$ & \soso \\
    $x/y$ & $10^2$ & \yes \\
    $1/\sqrt{x}$ & $10^2$ & \yes \\
    $\exp(x), \log(x)$ & $10^2$ & \yes \\
    \bottomrule
\end{tabular}
\end{table}

In this section, we give a high-level overview of our secure deep learning framework, and
how it builds upon our low-level computational modules (implemented from
scratch) to tackle challenges in secure computation. Table~\ref{table:basicops}
shows a non-exhaustive list of basic operations that is our computational
alphabet, with a rough estimation of their cost in MPC. The ``MPC Cost'' column
shows how an operation relates to integer multiplication in terms of
magnitude order because integer multiplication is the most basic operation that
involves communication. Most notably, comparison is not the basic operation
as in silicon computation.

\paragraph{Linear Operations}

Matrix multiplication is the workhorse of deep learning frameworks.
The forward and backward computation for both dense layers and convolutional
layers are implemented as matrix multiplication, which in turn is based on
dot products.
A particular challenge in secure computation is to compute a number of outputs
in parallel to save communication rounds. We overcome this challenge by
having a dedicated infrastructure that computes all dot products for a matrix
multiplication in \emph{a single batch of communication}, thus reducing the
number of communication rounds.
Another challenge is the accumulated error when evaluating matrix multiplication
using quantized numbers. The following Section~\ref{sec:error}
presents a careful analysis on our solution to this potential issue.
Our secure computation implementation uses efficient dot products
when available (\appref{sec:mpc}) or efficient use of SIMD-style homomorphic
encryption (\appref{sec:he}).

\paragraph{Non-linear Operations}

A major part of the non-linear computation is the non-linear neural network layers.
Both rectified linear unit (ReLU; \citealt{relu})
and max pooling are based on comparison followed by oblivious selection.
For example, $\rl(x) \defeq \max(x,0) = ( x>0~?~x~:~0 )$,
where the \texttt{C} expression $(\mathrm{condition}~?~a~:~b)$
outputs $a$ if $\mathrm{condition}$ is true and outputs $b$ if otherwise.
In secure computation, it saves communication
rounds if the process uses a balanced tree rather than iterating over
all input values of one maximum computation. For example,
two-dimensional max pooling over a 2x2 window requires computing the
maximum of four values. We compute the maximum of two pairs of values
followed by the maximum of the two results.
Our softmax and sigmoid activation functions are implemented
using the exponentiation algorithm in section~\ref{sec:exp}.
Another source of non-linear computation is the \emph{inverse square root
$1/\sqrt{x}$}, which appears in batch normalization~\citep{bn},
our secure version of Adam~\citep{adam} and AMSGrad~\citep{amsgrad} optimizers,
and Glorot initialization~\citep{glorot2010understanding} of the neural network weights.
Our novel implementation of $1/\sqrt{x}$ (explained in \switchref{sec:isqrt}) is
more efficient than state-of-the-art alternatives~\citep{ACNS:AlySma19,inv-sqrt}.
It helps to avoid numerical division, which is much more expensive than
multiplication.
Taking Adam as an example, the increment of the parameter $\theta_i$
in each learning step is altered
to $\gamma g_i / \sqrt{v_i + \varepsilon}$ ($\gamma$: learning rate;
$g_i$ and $v_i$: first and second moments of the gradient; $\varepsilon=10^{-8}$)
as compared to $\gamma g_i/\left(\sqrt{v_i} + \varepsilon\right)$
in \citeauthor{adam}'s original algorithm, so that we only need to
take the inverse square root of $v_i + \varepsilon$, then multiply
by $\gamma$ and $g_i$.

\paragraph{Random Operations}

Our secure framework supports random initialization of the network
weights~\cite{glorot2010understanding}, random rounding, random
shuffling of the training samples, and Dropout~\citep{dropout}.
Our implementation relies on uniform random samplers
\citep{USENIX:DalEscKel21,C:EGKRS20} based on mixed-circuit computation.
Random rounding and Dropout require to draw samples from Bernoulli random variables.
By a reparameterization trick, a sample $x$ from the distribution
$\mathrm{Prob}(x=a)=p$, $\mathrm{Prob}(x=b)=1-p$
($\mathrm{Prob}(\cdot)$: probability of statement being true)
can be expressed as $x=b+(a-b)\floor{p+r}$, where $r\sim\uniform(0,1)$ (uniform
distribution on $[0,1]$).

 \section{An Analysis of Probabilistic Rounding}\label{sec:error}

In this section, we analyze the rounding error of our secure quantized
representation for matrix multiplication. We show that our rounding algorithm
\ding{192} is unbiased; \ding{193} has acceptable error;
\ding{194} is potentially helpful to training.
Fixed point numbers~\citep{fixedpointcnn} and stochastic
rounding~\citep{stochasticrounding} have been applied to deep learning.
The rounding error analysis presented here is potentially useful to similar quantized representations.

Recall that any $x\in\Re$ is stored as an integer (equivalently, a fixed
point number) $Q^f(x)\defeq\round{x\cdot2^f}$.
As the maximum error of the nearest rounding $\round{\cdot}$
is $1/2$, we have
$\vert x-Q^f(x)2^{-f}\vert
=
2^{-f} \vert x\cdot2^f-Q^f(x)\vert
\le
2^{-f-1}=\epsilon/2$,
where $\epsilon\defeq{}2^{-f}$ is the smallest positive number we can represent.
Our multiplication algorithm outputs $R^f(xy)$, the integer representation of
the product $xy$. By definition,
\begin{equation}\label{eq:round}
\begin{array}{c}
R^f(xy) \defeq \floor{\mu} + b,\\[3pt]
\mu     \defeq Q^f(x) Q^f(y) 2^{-f},
\quad
b       \sim \bernoulli\left(\{\mu\}\right),
\end{array}
\end{equation}
where $\{\mu\}$ is the fractional part of $\mu$ ($0\le\{\mu\}<1$),
the random variable $b\in\{0,1\}$ means rounding down or up,
and $\bernoulli(\cdot)$ denotes a Bernoulli distribution with given mean.
Notice that the input $Q^f(x)$ and $Q^f(y)$ and the output $R^f(xy)$ are integers.
They all have to be multiplied by $2^{-f}$ to get the underlying fixed point number.
Eq.~(\ref{eq:round}) can be computed purely based on integer operations
(see \appref{sec:blocks}).
The Bernoulli random variable $b$ in Eq.~(\ref{eq:round}) is sampled
independently for each numerical product we compute.

Given two real matrices $A_{m\times{n}}$ and $B_{n\times{p}}$
represented by $Q^f(A)$ and $Q^f(B)$,
where $Q^f(\cdot)$ is elementwisely applied to the matrix entries,
the integer representation of their product is
$R^f(AB)\defeq\left(\sum_{l}R^f(A_{il}B_{lj})\right)_{m\times{p}}$.
Its correctness is guaranteed in the following proposition.
\begin{proposition}\label{thm:unbiased}
    $\forall{A}\in\Re^{m\times{n}}$, $\forall{B}\in\Re^{n\times{p}}$,
    $E(R^f(AB)) = 2^{-f} Q^f(A) Q^f(B)$,
    where $E(\cdot)$ denotes the expectation with respect to the
    random rounding in Eq.~(\ref{eq:round}).
\end{proposition}
The expectation of the random matrix $R^f(AB)$
is exactly the matrix product of $Q^f(A)$ and $Q^f(B)$ up constant scaling.
Hence, our matrix multiplication is \emph{unbiased}.
Product of fixed point numbers can also be based on nearest rounding
by replacing Eq.~(\ref{eq:round}) with $\tilde{R}^f(xy)\defeq\round{Q^f(x)Q^f(y)2^{-f}}$.
As a deterministic estimation, it is unbiased only if the rounding error is
absolutely zero, which is not true in general. Therefore, matrix product by nearest rounding is \emph{biased}.

We bound the worst case error during matrix multiplication as follows.
\begin{proposition}\label{thm:matbound}
Assume $A\in\Re^{m\times{n}}$, $B\in\Re^{n\times{p}}$, and the absolute value
of all entries are bounded by $2^k$ ($k\ge0$). Then,
$\Vert R^f(A B) - 2^{f}AB \Vert <
\sqrt{mp}\cdot{}n\cdot{}\left( 2^{k}+1 + \epsilon/4 \right)$,
where $\epsilon\defeq2^{-f}$, and $\Vert\cdot\Vert$ is the Frobenius norm.
\end{proposition}
In Proposition~\ref{thm:matbound}, $k$ corresponds to the number of bits to present the
integer part of any $x\in\Re$.
To get some intuition, we have the trivial bound $\Vert{AB}\Vert\le
\sqrt{mp}\cdot{}n\cdot2^{k+1}$. By Proposition~\ref{thm:matbound}, the computational error
$\Vert 2^{-f}R^f(A B) - AB \Vert$ has a smaller order of magnitude,
as it scales with $\epsilon=2^{-f}$. The precision $f$ should be high enough to
avoid a large error. Let us remark that nearest rounding has similar
deterministic error bounds (omitted here).

The following proposition shows our rounding in eq.~(\ref{eq:round}) satisfies
another \emph{probabilistic bound}, which improves over the above worst case bound.
\begin{proposition}\label{thm:std}
    Let $A\in\Re^{m\times{n}}$, $B\in\Re^{n\times{p}}$.
    With probability at least $1-\frac{1}{4\iota^2}$, the following is true
    \[
    \Vert R^f(AB) - 2^{-f}Q^f(A)Q^f(B) \Vert \le \iota \sqrt{mnp}.
    \]
\end{proposition}
By Proposition~\ref{thm:unbiased}, $2^{-f}Q^f(A)Q^f(B)$ is the expectation of
$R^f(AB)$.
Therefore the first term is the ``distance'' between the random matrix $R^f(AB)$ and its mean.
One can roughly approximate $2^{-f}Q^f(A)Q^f(B) \approx 2^fAB$
to compare Proposition~\ref{thm:std} with Proposition~\ref{thm:matbound}.
The probabilistic bound scales the deterministic bound by a factor of
$2^{-k}/{\sqrt{n}}$. That means our matrix multiplication is potentially more
accurate than alternative implementations based on nearest rounding,
especially for large matrices ($n$ is large).

One can write Dropout~\citep{dropout} in a similar way to eq.~(\ref{eq:round}):
$\mathrm{Dropout}(A) = A \odot B$, $b_{ij} \sim \bernoulli(\tau)$,
where $\odot$ is Hadamard product, and $\tau>0$ is a dropout rate.
Our probabilistic rounding is naturally equipped with noise injection,
which is helpful to avoid bad local optima and over-fitting.
Based on our observations, one can achieve better classification accuracy
than nearest rounding in general.

\section{Implementation and Benchmarks}
\label{sec:implementation}

We build our implementation on MP-SPDZ by
\citet{CCS:Keller20}. MP-SPDZ not only implements a range of MPC
protocols, but also comes with a high-level library containing
some of the building blocks in \appref{sec:blocks}. We have added the
exponentiation in Section \ref{sec:exp}, the inverse square
root computation in \switchref{sec:isqrt}, and all machine learning
building blocks.

MP-SPDZ allows implementing the computation in \texttt{Python} code, which is
then compiled into a specific bytecode. This code can be executed by a
virtual machine performing the actual secure computation. The process
allows optimizing the computation in the context of MPC.

The framework also features an emulator that executes the
exact computation in the clear that could be done securely. This
allowed us collecting
the accuracy figures in the next section at a lower cost.
Our implementation is licensed under a BSD-style license.

 \subsection{MNIST Classification}\label{sec:mnist}

For a concrete measurement of accuracy and running times, we
train a multi-class classifier\footnote{Scripts are available at
  \url{https://github.com/csiro-mlai/deep-mpc}.}
for the widely-used MNIST dataset~\citep{lecun2010mnist}.
We work mainly with the models used by
\citet{PoPETS:WagGupCha19} with secure computation, and we reuse
their numbering (A--D). The models contain up to four
layers. Network C is a convolutional neural network that appeared in
the seminal work of \citet{lecun1998gradient} whereas the others are
simpler networks used by related literature on secure
computation such as \citet{SP:MohZha17}, \citet{CCS:LJLA17}, and
\citet{ASIACCS:RWTSSK18}. See
Figures~\ref{fig:net-a}--\ref{fig:net-d} in the Appendix
for the exact network structures.

\begin{figure}[thb!]
  \centering
  \pgfplotsset{every axis legend/.append style = {legend pos = south west}}
\centering
\scalebox{.8}{\begin{tikzpicture}
\begin{axis}[
ylabel = training loss,
  xlabel = \#epochs,
ymode = log,
  ytick pos = left,
legend style = {cells={anchor=west}},
  legend entries = {
    {$f=16$, prob.},
    {$f=32$, prob.},
    {$f=64$, prob.},
    {$f=16$, nearest},
    {$f=32$, nearest},
  },
  ]
  \addplot[colour1] table[x=epoch, y=loss, mark=none]{data/trunc_pr};
  \addplot[colour2,densely dashed] table[x=epoch, y=loss, mark=none]{data/trunc_pr-f32-k63};
  \addplot[colour3,densely dashdotted] table[x=epoch, y=loss, mark=none]{data/trunc_pr-f64-k127};
  \addplot[colour4,densely dotted] table[x=epoch, y=loss, mark=none]{data/nearest};
  \addplot[colour5,densely dashdotdotted] table[x=epoch, y=loss, mark=none]{data/nearest-f32-k63};
  \addplot [name path=upper,draw=none] table[x=epoch,y=loss_max] {data/trunc_pr};
  \addplot [name path=lower,draw=none] table[x=epoch,y=loss_min] {data/trunc_pr};
  \addplot [colour1!20] fill between[of=upper and lower];
  \addplot [name path=upper,draw=none] table[x=epoch,y=loss_max] {data/trunc_pr-f32-k63};
  \addplot [name path=lower,draw=none] table[x=epoch,y=loss_min] {data/trunc_pr-f32-k63};
  \addplot [colour2!20] fill between[of=upper and lower];
  \addplot [name path=upper,draw=none] table[x=epoch,y=loss_max] {data/nearest-f32-k63};
  \addplot [name path=lower,draw=none] table[x=epoch,y=loss_min] {data/nearest-f32-k63};
  \addplot [colour5!20] fill between[of=upper and lower];
\end{axis}
\begin{axis}[
  ylabel = testing error,
  ylabel near ticks,
  ytick pos = right,
  y dir = reverse,
yticklabel style={/pgf/number format/.cd,fixed,precision=2},
  scaled y ticks = false,
  ]
  \addplot[colour1] table[x=epoch, y=error, mark=none]{data/trunc_pr};
  \addplot[colour2,densely dashed] table[x=epoch, y=error, mark=none]{data/trunc_pr-f32-k63};
  \addplot[colour3,densely dashdotted] table[x=epoch, y=error, mark=none]{data/trunc_pr-f64-k127};
  \addplot[colour4,densely dotted] table[x=epoch, y=error, mark=none]{data/nearest};
  \addplot[colour5,densely dashdotdotted] table[x=epoch, y=error, mark=none]{data/nearest-f32-k63};
  \addplot [name path=upper,draw=none] table[x=epoch,y=error_max] {data/trunc_pr};
  \addplot [name path=lower,draw=none] table[x=epoch,y=error_min] {data/trunc_pr};
  \addplot [colour1!20] fill between[of=upper and lower];
  \addplot [name path=upper,draw=none] table[x=epoch,y=error_max] {data/trunc_pr-f32-k63};
  \addplot [name path=lower,draw=none] table[x=epoch,y=error_min] {data/trunc_pr-f32-k63};
  \addplot [colour2!20] fill between[of=upper and lower];
  \addplot [name path=upper,draw=none] table[x=epoch,y=error_max] {data/nearest-f32-k63};
  \addplot [name path=lower,draw=none] table[x=epoch,y=error_min] {data/nearest-f32-k63};
  \addplot [colour5!20] fill between[of=upper and lower];
\end{axis}
\end{tikzpicture}
}
\caption{Loss and accuracy for LeNet and various precision options when
  running SGD with a learning rate of 0.01. Interval only available for
  $f=32$ and $f=16$ with probabilistic rounding. \label{fig:c}}

 \vspace{.8em}
\centering
\scalebox{.8}{
\begin{tikzpicture}
\begin{axis}[
ylabel = training loss,
  xlabel = \#epochs,
ymode = log,
  ytick pos = left,
legend style = {cells={anchor=west}},
  legend entries = {
    {SGD, $\gamma=0.01$},
    {SGD, $\gamma=0.05$},
    {Adam, $\gamma=0.001$},
    {AMSGrad, $\gamma=0.001$},
    {AMSGrad, $\gamma=0.005$},
    {AMSGrad, $\gamma=0.001$, Dropout},
    {Quotient AMSGrad, $\gamma=0.001$},
  },
  ]
  \addplot[colour1] table[x=epoch, y=loss, mark=none]{data/trunc_pr};
  \addplot[colour2] table[x=epoch, y=loss, mark=none]{data/rate.1-trunc_pr};
  \addplot[colour6,densely dotted] table[x=epoch, y=loss, mark=none]{data/adamapprox-rate.001-trunc_pr};
  \addplot[colour4,densely dashdotted] table[x=epoch, y=loss, mark=none]{data/amsgrad-rate.001-trunc_pr};
  \addplot[colour3,densely dashdotted] table[x=epoch, y=loss, mark=none]{data/amsgrad-rate.005-trunc_pr};
  \addplot[colour8,densely dashdotted] table[x=epoch, y=loss, mark=none]{data/dropout1-amsgrad-rate.001-trunc_pr};
  \addplot[colour7,densely dashdotdotted] table[x=epoch, y=loss, mark=none]{data/quotient-rate.001-trunc_pr};
  \addplot [name path=upper,draw=none] table[x=epoch,y=loss_max] {data/amsgrad-rate.001-trunc_pr};
  \addplot [name path=lower,draw=none] table[x=epoch,y=loss_min] {data/amsgrad-rate.001-trunc_pr};
  \addplot [colour4!20] fill between[of=upper and lower];
  \addplot [name path=upper,draw=none] table[x=epoch,y=loss_max] {data/dropout1-amsgrad-rate.001-trunc_pr};
  \addplot [name path=lower,draw=none] table[x=epoch,y=loss_min] {data/dropout1-amsgrad-rate.001-trunc_pr};
  \addplot [colour8!20] fill between[of=upper and lower];
\end{axis}
\begin{axis}[
  ylabel = testing error,
  ylabel near ticks,
  ytick pos = right,
  y dir = reverse,
scaled y ticks = false,
  yticklabel style={/pgf/number format/.cd,fixed,precision=2},
  ]
  \addplot[colour1] table[x=epoch, y=error, mark=none]{data/trunc_pr};
  \addplot[colour2] table[x=epoch, y=error, mark=none]{data/rate.1-trunc_pr};
  \addplot[colour6,densely dotted] table[x=epoch, y=error, mark=none]{data/adamapprox-rate.001-trunc_pr};
  \addplot[colour4,densely dashdotted] table[x=epoch, y=error, mark=none]{data/amsgrad-rate.001-trunc_pr};
  \addplot[colour3,densely dashdotted] table[x=epoch, y=error, mark=none]{data/amsgrad-rate.005-trunc_pr};
  \addplot[colour8,densely dashdotted] table[x=epoch, y=error, mark=none]{data/dropout1-amsgrad-rate.001-trunc_pr};
  \addplot[colour7,densely dashdotdotted] table[x=epoch, y=error, mark=none]{data/quotient-rate.001-trunc_pr};
  \addplot [name path=upper,draw=none] table[x=epoch,y=error_max] {data/amsgrad-rate.001-trunc_pr};
  \addplot [name path=lower,draw=none] table[x=epoch,y=error_min] {data/amsgrad-rate.001-trunc_pr};
  \addplot [colour4!20] fill between[of=upper and lower];
  \addplot [name path=upper,draw=none] table[x=epoch,y=error_max] {data/dropout1-amsgrad-rate.001-trunc_pr};
  \addplot [name path=lower,draw=none] table[x=epoch,y=error_min] {data/dropout1-amsgrad-rate.001-trunc_pr};
  \addplot [colour8!20] fill between[of=upper and lower];
\end{axis}
\end{tikzpicture}
}
\caption{Loss and accuracy for LeNet with various optimizer options, $f=16$,
  and probabilistic truncation. $\gamma$ is the learning rate. Interval only
  available for AMSGrad with $\gamma=0.0001$. \label{fig:c-optimizers}}

 \vspace{.8em}
\centering
\scalebox{.8}{\begin{tikzpicture}
\begin{axis}[
ylabel = training loss,
  xlabel = \#epochs,
ymode = log,
  ytick pos = left,
legend style = {cells={anchor=west}},
  legend entries = {
    {secure, SGD, $\gamma=0.01$},
    {secure, AMSGrad, $\gamma=0.001$},
    {cleartext, SGD, $\gamma=0.01$},
    {cleartext, AMSGrad, $\gamma=0.001$},
  },
  ]
  \addplot[colour1] table[x=epoch, y=loss, mark=none]{data/trunc_pr};
  \addplot[colour2,densely dotted] table[x=epoch, y=loss, mark=none]{data/amsgrad-rate.001-trunc_pr};
  \addplot[colour3,densely dashdotted] table[x index=0, y index=2, mark=none]{data/log-C-50-.01};
  \addplot[colour5,densely dashdotdotted] table[x index=0, y index=2, mark=none]{data/log-C-50-amsgrad.001};
\end{axis}
\begin{axis}[
  ylabel = testing error,
  ylabel near ticks,
  ytick pos = right,
  y dir = reverse,
scaled y ticks = false,
  yticklabel style={/pgf/number format/.cd,fixed,precision=2},
  ]
  \addplot[colour1] table[x=epoch, y=error, mark=none]{data/trunc_pr};
  \addplot[colour2,densely dotted] table[x=epoch, y=error, mark=none]{data/amsgrad-rate.001-trunc_pr};
  \addplot[colour3,densely dashdotted] table[x index=0, y expr=1-\thisrowno{8}, mark=none]{data/log-C-50-.01};
  \addplot[colour5,densely dashdotdotted] table[x index=0, y expr=1-\thisrowno{8}, mark=none]{data/log-C-50-amsgrad.001};
\end{axis}
\end{tikzpicture}
}
\caption{Comparison of cleartext training and secure training for LeNet with $f=16$ and probabilistic truncation. $\gamma$ is the learning rate.\label{fig:comp}}
 \end{figure}

Figure~\ref{fig:c} shows the learning curves for various quantization
precisions and two rounding options, namely, nearest and probabilistic rounding.
We use SGD with learning rate 0.01, batch size 128, and the usual MNIST training/test split.
Most configurations perform similarly except for 16-bit precision
with nearest rounding which gives worse results.
We run the best-performing configurations several times.
The range of the performance scores is indicated by the shaded area.
For the rest of the paper, we focus on $f=16$ with probabilistic
rounding because it offers the best overall performance in terms of accuracy
and efficiency.

The choice of $f=16$ (and $k=31$) is informed on one hand by the fact that it
is close to the lower limit for reasonable results. The works listed in
Table~\ref{table:mpc} use values in the range 13--20, and we found that $f=8$
leads to divergence. On the other hand,
Figure~\ref{fig:c} shows that increasing the precision does not lead to an
increase in accuracy. Finally, the division algorithm by \citet{FC:CatSax10}
requires that $k$ is roughly twice $f$, and that $k$ is half the domain size
(strictly less than half the domain size for efficiency). A domain size of 64
is natural given the 64-bit word size of common hardware. The choice of $f$
and $k$ follows from these constraints.
Learning rate and minibatch size are set empirically based on
satisfactory performance in plaintext training.

Figure \ref{fig:c-optimizers} reports the results with a variety of
optimizers. AMSGrad~\citep{amsgrad} stands out in terms of convergence and final
accuracy.
\citet{CCS:ASKG19} suggest using normalized gradients in AMSGrad
for training binary neural networks to improve performance.
In our experiments using quantized weights with higher precision,
similar improvements are not observed.

Furthermore, Figure \ref{fig:c-optimizers} also shows that using a
Dropout layer as described in Figure \ref{fig:net-c} in
\switchref{sec:model-code} only slightly
improves the performance.
This is possibly due to the fact that the reduced precision and probabilistic
rounding already prevent overfitting to some degree.

\begin{table*}
\captionsetup{width=.96\textwidth}
  {
  
\centering
  \tcaption{Benchmarks in the three-party LAN
    setting with one corruption. Accuracy N/A means that
    the accuracy figures were not given or computed in a way that does
    not reflect the secure computation. ($^*$) \citet{PoPETS:WTBKMR21}
    only implemented their online phase.}
  \label{table:mpc}
  \begin{tabular}{llrrrr}
    \toprule
    Network && Epoch time (s) & Comm. per epoch (GB) & Acc. (\# epochs) & Precision ($f$) \\
    \midrule
    \multirow{5}{*}{A} & \citet{CCS:MohRin18} & 180 & N/A & 94.0\% (15) & N/A \\
            & \citet{PoPETS:WagGupCha19} & 247 & N/A & N/A & 13 \\
            & \citet{PoPETS:WTBKMR21}$^*$ & 41 & 3 & N/A & 13 \\
            & Ours (SGD) & 31 & 26 & 97.8\% (15) & 16 \\
            & Ours (AMSGrad) & 88 & 139 & 98.1\% (15) & 16 \\
    \midrule
    \multirow{4}{*}{B} & \citet{PoPETS:WagGupCha19} & 4,176 & N/A & N/A & 13 \\
            & \citet{PoPETS:WTBKMR21}$^*$ & 891 & 108 & N/A & 13 \\
            & Ours (SGD) & 144 & 201 & 98.0\% (15) & 16 \\
            & Ours (AMSGrad) & 187 & 234 & 98.6\% (15) & 16 \\
    \midrule
    \multirow{6}{*}{C} & \citet{PoPETS:WagGupCha19} & 7,188 & N/A & N/A & 13 \\
            & \citet{PoPETS:WTBKMR21}$^*$ & 1,412 & 162 & N/A & 13 \\
            & \citet{tan2021cryptgpu} & 1,036 & 534 & 94.0\% (5) & 20  \\
            & \citet{crypten} & 10,940 & N/A & 96.7\% (4) & 16 \\
            & Ours (SGD) & 343 & 352 & 98.5\% (5) & 16 \\
            & Ours (AMSGrad) & 513 & 765 & 99.0\% (5) & 16 \\
    \midrule
    \multirow{4}{*}{D} & \citet{CCS:MohRin18} & 234 & N/A & N/A & N/A \\
            & \citet{PoPETS:WTBKMR21}$^*$ & 101 & 11 & N/A & 13 \\
            & Ours (SGD) & 41 & 41 & 98.1\% (15) & 16 \\
            & Ours (AMSGrad) & 95 & 137 & 98.5\% (15) & 16 \\
    \bottomrule
  \end{tabular}

   \vspace{1em}
  
\centering
  \tcaption{Benchmarks in the two-party LAN
    setting. In the column ``Epoch time (s)'', two numbers
    refer to online and offline time. Accuracy N/A means that
    the accuracy figures were not given or computed in a way that does
    not reflect the secure computation. Figures for Network C are
    estimates based on ten batch iterations.}
  \label{table:mnist-2pc}
  \begin{tabular}{llrrrr}
    \toprule
    Network && Epoch time (s) & Comm. per epoch (GB) & Acc. (\# epochs) & Precision ($f$) \\
    \midrule
    \multirow{4}{*}{A} & \citet{SP:MohZha17} & 283/19,333 & N/A & 93.4\% (15) & 13 \\
            & \citet{CCS:ASKG19} & 31,392 & N/A & 95.0\% (10) & N/A \\
            & Ours (SGD) & 3,741 & 1,128 & 97.8\% (15) & 16 \\
            & Ours (AMSGrad) & 5,688 & 4,984 & 98.1\% (15) & 16 \\
    \midrule
    \multirow{2}{*}{D} & Ours (SGD) & 141,541 & 25,604  & 98.1\% (15) & 16 \\
            & Ours (AMSGrad) & 196,745 & 51,770 & 98.5\% (15) & 16 \\
    \bottomrule
  \end{tabular}

   \vspace{1em}
  \centering
  \tcaption{Benchmarks for Network C (LeNet) with AMSGrad across a
    range of security models in the LAN setting. Figures are mostly
    estimates based on ten batch iterations.}
  \label{table:sec-models}
  \begin{tabular}{lcccrrr}
    \toprule
    Protocol & \# parties & \# corruptions & Malicious & Time per epoch (s) & Comm. per epoch (GB) \\
    \midrule
    Appendix \ref{sec:he} & 2 & 1 & \tock & 196,745 & 51,770 \\
    Appendix \ref{sec:mpc} & 3 & 1 & \tock & 513 & 765 \\
    \citet{USENIX:DalEscKel21} & 3 & 1 & \tick & 4,961 & 9,101 \\
    Appendix \ref{sec:he} & 3 & 2 & \tock & 357,214 & 271,595 \\
    \citet{USENIX:DalEscKel21} & 4 & 1 & \tick & 1,175 & 2,945 \\
    Appendix \ref{sec:dealer} & 10 & 1/8 & \tock & 29,078 & 99,775 \\
    \citet{C:GLOPS21} & 10 & 4 & \tock & 129,667 & 434,138 \\
    \appref{sec:he} & 10 & 9 & \tock & 2,833,641 & 13,875,834 \\
    \bottomrule
  \end{tabular}

   }
\end{table*}

\paragraph{Resources}

We run the emulator on AWS \texttt{c5.9xlarge} instances. One epoch
takes a few seconds to several minutes depending on the model that is being trained.
For all our experiments, we used a few weeks of computational time
including experiments not presented here.

\subsubsection{Secure computation}

In order to verify our emulation results, we run LeNet with
precision $f=16$ and probabilistic rounding in our actual multi-party
computation protocol. We could verify that it converges on 99.2\%
accuracy at 25 epochs, taking 3.6 hours (or one hour for 99\%
accuracy). Tables~\ref{table:mpc} and \ref{table:mnist-2pc} compare
our results to previous works
in a LAN setting.

Note that \citet{PoPETS:WagGupCha19} and
\citet{PoPETS:WTBKMR21} give accuracy figures. The authors have told us
in personal communication that their figures do not reflect the secure
computation.
This is related to the fact that they do not implement exponentiation
and thus softmax but use the ReLU-based alternative, which has been
found to perform poorly by \citet{keller2020effectiveness}. The same
holds for \citet{CCS:MohRin18} and \citet{SP:MohZha17}.
\citet{CCS:ASKG19} instead use ternary weights, which explains their
reduced accuracy.

\citet{tan2021cryptgpu} rely entirely on Taylor series approximation
for exponentiation whereas we use approximation only for the fractional
part of the input (Algorithm \ref{alg:our-exp}). This might explain their
lower accuracy.

The figures for CrypTen \citep{crypten} were obtained by running an
adaption of their \verb+mpc_autograd_cnn+ example to the full
LeNet with SGD and learning rate 0.01. We found that it would eventually
diverge even with a relatively low learning rate of 0.001. The given
accuracy is the highest we achieved before divergence.
We estimate that this is due to the approximations used in CrypTen. It
use the following approximation for the exponential function:
\[ (1 + x/2^n)^{2^n}. \]
According to the CrypTen
code\footnote{\url{https://github.com/facebookresearch/CrypTen}},
$n$ is set to 9. Furthermore, they use a fixed-point precision of 16.
This leads to a relatively low precision. For $x=-4$, their approximation
is $0.018030$ when rounding to the nearest multiple of $2^{-16}$ at every
step. In contrast, the correct value is $0.018315$, a relative error of more
than one percent. In comparison, our solution produces values in the range
$[0.018310, 0.018325]$, a relative error of less than $10^{-3}$ and an
absolute error of less than the representation step of $2^{-16}$.

There are a number of factors that make it hard to compare the
performance. \citet{PoPETS:WTBKMR21} have only implemented their
online phase, which makes their figures incomplete. The implementation
of \citet{crypten} is purely in \texttt{Python} whereas all others use
\texttt{C/C++}. Finally, we rely on figures obtained using different platforms
except for \citet{crypten}.
In general, it is likely that MP-SPDZ benefits from being an established
framework that has undergone more work than other code bases.

\paragraph{Further security models}
\label{sec:sec-models}

MP-SPDZ supports a wide range of security models, that is, choices for
the number of (corrupted) parties and the nature of their corruption,
i.e., whether they are assumed to follow the protocol or not. Table
\ref{table:sec-models} shows our results across a range of security
models.

\subsubsection{Comparison to Cleartext Training}

Figure \ref{fig:comp} compares the performance of our secure training
with cleartext training in TensorFlow. It shows that secure training
performs only slightly worse with the same optimizer.

 \subsection{CIFAR-10 Classification}
\label{sec:cifar10}

In order to highlight the generality of our approach, we have implemented training
for CIFAR-10 with a network by \citet{PoPETS:WTBKMR21}, which is based
on AlexNet~\citep{alexnet}.
See Figure \ref{fig:alex} in \appref{sec:model-code} for details.
Unlike the networks for MNIST, the network for CIFAR-10 uses batch
normalization~\citep{bn}. Table~\ref{table:cifar10} shows our results.
Our results are not comparable with \citet{PoPETS:WTBKMR21}
because they do not implement the full training.
Furthermore, \citet{tan2021cryptgpu} do not provide accuracy figures
for training from scratch, which makes it unclear how their less
accurate approximations perform in this setting.
Figure \ref{fig:comp-cifar10} in \appref{sec:more-figures} shows that
our secret training comes within
a few percentage points of cleartext training.
Similar figures for MPC training on CIFAR-10 are not widely seen in the
literature.

\begin{table}[h!]
  \centering
  \tcaption{Time (seconds) and communication (GB) per epoch, accuracy
    after ten epochs, and fixed-point precision for CIFAR-10 training
    in the three-party LAN setting with one
    corruption. \citet{tan2021cryptgpu} do not train from scratch,
    which is why we do include their accuracy
    figure. ($^*$)\citet{PoPETS:WTBKMR21} only implemented their
    online phase.}
  \label{table:cifar10}
  \begin{tabular}{lrrrr}
    \toprule
    Network & s/ep. & GB/ep. & Acc. & $f$ \\
    \midrule
    \citet{PoPETS:WTBKMR21}$^*$ & 3,156 & 80 & N/A & 13 \\
    \citet{tan2021cryptgpu} & 1,137 & 535 & N/A & 20 \\
    Ours (SGD) & 1,603 & 771 & 64.9\% & 16 \\
    Ours (Adam) & 2,431 & 3,317 & 64.7\% & 16 \\
    Ours (AMSGrad) & 2,473 & 3,285 & 63.6\% & 16 \\
    \bottomrule
  \end{tabular}
\end{table}

\section{Conclusions}
\label{sec:conclusions}

We implement deep neural network training purely in multi-party
computation, and we present extensive results
of convolutional neural networks on benchmark datasets.
We find that the low precision of MPC computation increases the
error slightly. We only consider one particular
implementation of division and
exponentiation, which are crucial to the learning process as part of
softmax.
Future work might consider different approximations of these
building blocks.

\bibliographystyle{icml2022}
\FloatBarrier

\newpage
\onecolumn
\captionsetup{skip=.5em,width=0.9\textwidth}
\appendix

\section{An Efficient Secure Three-Party Computation Protocol}
\label{sec:mpc}

There is a wide range of MPC protocols with a variety of security
properties (see \citet{CCS:Keller20} for an overview). In this section, we
focus on the setting of \emph{three-party computation with one semi-honest
corruption}. This means that, out of the three parties, two are expected
to behave honestly, i.e., they follow the protocol and keep their view of the
protocol secret, and one party is expected to follow the protocol but
might try to extract information from their view. The reason for choosing
this setting is that it allows for an efficient MPC protocol while still
providing secure outsourced computation.
The protocol does not easily generalize to any other setting.
However, protocols exist for any number of parties, see Keller
for an overview.
The concrete protocol we use
goes back to \citet{C:BenLei88} with further aspects by
\citet{CCS:AFLNO16}, \citet{CCS:MohRin18}, and \citet{ITC:EKOPPS20}.
We summarize the core protocol below. The mathematical building blocks
in the next section mostly use the aspects below.

\paragraph{Secret sharing}
The simplest variant of secure computation is only data-oblivious,
that is, the participants are aware of the nature of the computation
(addition, multiplication, etc.) but not the values being operated
on. In the context of machine learning, this means that they know the
hyper-parameters such as the layers but not the sample data or the
neural network weights.
All of these values  in our protocol are stored using replicated
secret sharing. A secret value $x$ is a represented as a random sum
$x = x_0 + x_1 + x_2$, and party $P_i$ holds $(x_{i-1}, x_{i+1})$
where the indices are computed modulo three. Clearly, each party is
missing one value to compute the sum. On the other hand, each pair of
parties hold all necessary to reconstruct the secret. For a uniformly
random generation of shares, the computation domain has to be
finite. Most commonly, this domain is defined by integer computation
modulo a number. We use $2^k$ for $k$ being a multiple of 64 as well
as 2 as the
moduli. The first case corresponds to an extension of 64-bit
arithmetic found on most  processors. We will
refer to the two settings as arithmetic and binary secret sharing
throughout the paper.

\paragraph{Input sharing}
The secret sharing scheme implies a protocol to share inputs where the
inputting party samples the shares and distributes them
accordingly. \citet{ITC:EKOPPS20} propose a more efficient
protocol where the inputting party only needs to send one value
instead of two pairs of values. If $P_i$ would like to input $x$,
$x_i$ is set to zero, and $x_{i-1}$ is generated with a pseudo-random
generator using a key previously shared between $P_i$ and
$P_{i+1}$. $P_i$ can compute $x_{i+1} = x - x_{i-1}$ and send it to
$P_{i-1}$. While the resulting secret sharing is not entirely random,
the fact that $P_i$ already knows $x$ makes randomizing $x_i$ obsolete.

\paragraph{Addition, subtraction, and scalar multiplication}
The commutative nature of addition allows to add secret sharings
without communication. More concretely, secret sharings $x = x_0 + x_1
+ x_2$ and $y = y_0 + y_1 + y_2$ imply the secret sharing $x + y = (x_0 +
y_0) + (x_1 + y_1) + (x_2 + y_2)$.
The same works for subtraction. Furthermore, the secret sharing
$x = x_0 + x_1 + x_2$ allows to compute $\lambda x = \lambda x_0 +
\lambda x_1 + \lambda x_2$ locally.

\paragraph{Multiplication}
The product of $x = x_0 + x_1 + x_2$ and $y = y_0 + y_1 + y_2$ is
\begin{align*}
  x \cdot y &= (x_0 + x_1 + x_2) \cdot (y_0 + y_1 + y_2) \\
  &= (x_0y_0 + x_0y_1 + x_1y_0)
    + (x_1y_1 + x_1y_2 + x_2y_1)
    + (x_2y_2 + x_2y_0 + x_0y_2).
\end{align*}
On the right-hand side (RHS),
each term in parentheses only contains shares known by one of the
parties. They can thus compute an additive secret sharing (one summand
per party) of the
product. However, every party only holding one share does not satisfy
the replication requirement for further multiplications. It is not
secure for every party to pass their value on to another party because
the summands are not distributed randomly. This can be fixed by
rerandomization: Let $xy = z_0 + z_1 + z_2$ where $z_i$ is known to
$P_i$. Every party $P_i$ computes $z_i' = z_i + r_{i,i+1} - r_{i-1,i}$
where $r_{i,i+1}$ is generated with a pseudo-random generator using a
key pre-shared between $P_i$ and $P_{i+1}$. The resulting sum $xy =
z_0' + z_1' + z_2'$ is pseudo-random, and it is thus secure for $P_i$
to send $z_i'$ to $P_{i+1}$ in order to create a replicated secret
sharing $((xy)_{i-1}, (xy)_{i+1}) = (z_i', z_{i-1}')$.

\paragraph{Domain conversion}

Recall that we use computation modulo $2^k$ for $k$ being a multiple
of 64 as well as 1. Given that the main operations are just addition
and multiplication in the respective domain, it is desirable to
compute integer arithmetic in the large domain but operations with a
straight-forward binary circuit modulo two. There has been a
long-running interest in this going back to least
\citet{kolesnikov2013systematic}. We mainly rely on the approach
proposed by \citet{CCS:MohRin18} and \citet{CCS:ABFKLO18}. Recall that
$x \in 2^k$ is shared as $x = x_0 + x_1 + x_2$. Now let
$\{x_0^{(i)}\}_{i=0}^{k-1}$ the bit decomposition of $x_0$, that is,
$x_0^{(i)} \in \{0,1\}$ and $x_0 = \sum_{i=0}^{k-1} x_0^{(i)} 2^i$. It
is self-evident that $x_0^{(i)} = x_0^{(i)} + 0 + 0$ is a valid secret
sharing modulo two (albeit not a random one). Furthermore, every party
holding $x_0$ can generate $x_0^{(i)}$. It is therefore possible for
the parties to generate a secret sharing modulo two of a
\emph{single share} modulo
$2^k$. Repeating this for all shares and computing the addition as
a binary circuit allow the parties to generate a secret sharing
modulo two from a secret sharing modulo $2^k$.  Conversion in the
other direction can be achieved using a similar technique or using
``daBits'' described by \citet{INDOCRYPT:RotWoo19}. In the following,
we will use the term mixed-circuit computation for any technique that
works over both computation domains.

\paragraph{Dot products}

Dot products are an essential building block of linear computation
such as matrix multiplication. In light of quantization, it is
possible to reduce the usage of truncation by deferring it to after the
summation. In other words, the dot product in the integer
representations is computed before truncating. This not only reduces
the truncation error, but also is more efficient because the truncation
is the most expensive part in quantized secure multiplication.
Similarly, our protocol allows to defer the communication needed for
multiplication. Let $\vec{x}$ and $\vec{y}$ be two vectors where the
elements are secret-shared, that is, $\{x^{(i)}\} = x_0^{(i)} + x_1^{(i)} +
x_2^{(i)}$ and similarly for $y^{(i)}$. The inner product then is
\begin{align*}
  \sum_i x^{(i)} \cdot y^{(i)}
  &= \sum_i (x_0^{(i)} + x_1^{(i)} + x_2^{(i)}) \cdot (y_0^{(i)} +
    y_1^{(i)} + y_2^{(i)}) \\
  &= \sum_i (x_0^{(i)}y_0^{(i)} + x_0^{(i)}y_1^{(i)} +
    x_1^{(i)}y_0^{(i)}) + \sum_i (x_1^{(i)}y_1^{(i)} +
    x_1^{(i)}y_2^{(i)} + x_2^{(i)}y_1^{(i)}) \\
  &\quad + \sum_i
    (x_2^{(i)}y_2^{(i)} + x_2^{(i)}y_0^{(i)} + x_0^{(i)}y_2^{(i)}).
\end{align*}
The three sums in the last term can be computed locally by one party
each before applying the same protocol as for a single multiplication.

\paragraph{Comparisons}

Arithmetic secret sharing does not allow to access the individual
bits directly. It is therefore not straightforward to compute
comparisons such as ``less than''. There is a long line of literature
on how to achieve this going back to at least
\citet{TCC:DFKNT06}. More recently, most attention has been given to
combining the power of arithmetic and binary secret sharing in order
to combine the best of worlds. One possibility to do so is to plainly
convert to the binary domain and compute the comparison circuit there.
In our concrete implementation, we use the more efficient approach
by \citet{CCS:MohRin18}. It
starts by taking the difference between the two inputs. Computing the
comparison then reduces to comparing to zero, which in turn is
equivalent to extracting the most significant bit as it indicates the
sign. The latter is achieved by converting the shares locally to bit-wise
sharing of the arithmetic shares, which sum up to the secret value. It
remains to compute the sum of the binary shares in order to come up
with the most significant bit.

\paragraph{Shifting and truncation}

\citet{USENIX:DalEscKel21} present an efficient implementation of the
deterministic truncation using mixed-circuit computation, and
\citet{PoPETS:DalEscKel20} present an efficient protocol for the
probabilistic truncation.

The probabilistic truncation involves the truncation of a randomized
value, that is the computation of $\floor{(x+r)/2^f}$ for a random
$f$-bit value $r$. It is easy to see that
\[ \floor{(x+r)/2^f} = \begin{cases}
    \floor{x/2^f} & \text{if }(x \bmod 2^f + r) < 2^f \\
    \floor{x/2^f} + 1 & \text{otherwise.}
  \end{cases} \]
Therefore, the larger $(x \bmod 2^f)$ is, the more
likely the latter condition is true.

 \section{An Efficient Multi-Party Computation Protocol Based on Homomorphic Encryption}
\label{sec:he}

While the protocol in the previous is very efficient, the
multiplication only works with an honest majority, that is, the number
of corrupted parties is strictly less than half of the total
number. Multiplications with more corruptions than that requires more
involved cryptographic schemes. In this section, we present a
semi-honest protocol based on homomorphic encryption that tolerates
$n-1$ corrupted parties out of $n$.

\paragraph{Secret sharing}
The simplest secret sharing is additive secret sharing, which can be
used in this setting. The share of party $i$ is a random element $x_i$
of the relevant domain such that the sum of all elements is the
secret.

\paragraph{Input sharing}
The canonical input sharing would be for the inputting party to
generate random shares such that they add up to the input and then
distribute the shares. Using a one-time setup, it is possible to share
without communication as shown in Algorithm \ref{alg:add-sharing}.  It
is easy to see that the resulting sharing is correct and that the
unknown shares are indistinguishable to random for an adversary
corrupting any number of parties.

\begin{algorithm}
\caption{Additive input sharing without communication.}
\label{alg:add-sharing}
\begin{algorithmic}[1]
  \Setup{PRG key $K_{ij}$ known only to parties $P_i$ and $P_j$}
  \Input{Party $P_i$ has input $x$.}
  \Output{\share{x}}
  \State Party $P_i$ sets $x_i \asn x - \sum_{j \ne i} \PRG_{K_{ij}}()$.
  \State For all $j \ne i$, party $P_j$ sets $x_j \asn \PRG_{K_{ij}}()$.
\end{algorithmic}
\end{algorithm}

\paragraph{Addition, subtraction, and scalar multiplication}
As with any linear secret sharing scheme, these operation can
trivially be computed locally.

\paragraph{Multiplication}
We use Beaver's technique \citep{C:Beaver91b}. Assume that $a$ and $b$ are
random numbers in the domain, and $c=ab$. Then, for any $x, y$, it
holds that
\begin{align*}
  xy &= (x + a - a) \cdot (y + b - b) \\
  &= (x + a) \cdot (y + b) - (y +
    b) \cdot a - (x + a) \cdot b + c.
\end{align*}
As $a$ and $b$ and random, they can be used to used to mask $x$ and
$y$, respectively. Once $x+a$ and $y+b$ have been revealed, the
operation becomes linear in $a$ and $b$:
\[ [xy] = (x + a) \cdot (y + b) - (y +
  b) \cdot [a] - (x + a) \cdot [b] + [c]
\]

It remains to produce a fresh triple $(a, b, c)$ for every
multiplication. \citet{EC:KelPasRot18} show how to do this using
semi-homomorphic encryption based on Learning With Errors (LWE), a
widely used cryptographic assumption. They make optimal use of the
SIMD (single instruction, multiple data) nature of LWE-based
encryption. Their protocol involves pair-wise communication between
parties, which scale quadratically in the number of parties. In the
following section, we propose a protocol that scales better, which is
inspired by \citet{EC:CraDamNie01}.

\paragraph{Matrix multiplication}

\citet{matrix-triples} have extended Beaver's technique to matrix
multiplication by replacing $x, a$ with $m \times n$ and $y, b$ with
$n \times p$ matrices. Using the diagonal packing by
\citet{C:HalSho14} allows to implement the matrix triple generation
with $O(m (n + l))$ communication instead of the na\"{i}ve
$O(mnl)$. Unlike Halevi and Shoup, we only use the semi-homomorphic
encryption, that is, only public-private multiplications. This avoids
the need for cycling or shifting because we can simply generate the
necessary cleartexts as needed.

\paragraph{Domain conversion}

daBits \citep{INDOCRYPT:RotWoo19} and edaBits \citep{C:EGKRS20} are a generic
method to convert between integer and binary computation. For the
special case of semi-honest two-party computation,
\citet{NDSS:DemSchZoh15} have presented a more efficient protocol
similar to the one in the previous section.

\paragraph{Comparison, shifting, and truncation}

\citet{C:EGKRS20} have shown how to use edaBits to implement these
computations. Essentially, they make minimal use of binary computation
before switching back to integer computation so that the result is
available for further computation in the integer domain.

\subsection{Linear-cost Triple Generation with Semi-homomorphic
  Encryption}

Assume an encryption system that allows to multiply a ciphertext and a
plaintext, that is, $\Dec(\Enc(\va) \cdot \vb = \va * \vb$ where $\va$
and $\vb$ are vectors and $*$ denotes the Schur product. LWE allows
the efficient creation of such a system. Furthermore, assume a
distributed key setup as used by \citet{C:DPSZ12}. Since we only want
to achieve semi-honest security, it is straight-forward to generate
without using any protocol to check the correctness. Algorithm
\ref{alg:triple} show the details of our protocol.

\begin{algorithm}
\caption{Triple Generation with distributed semi-homomorphic encryption}
\label{alg:triple}
\begin{algorithmic}[1]
  \Input{Distributed key setup}
  \Output{Random multiplication triples}
  \State Each party $P_i$ samples $\va_i$ and send $\Enc(\va_i)$ to $P_1$.
  \State $P_1$ computes $C_a = \sum_i \Enc(\va_i)$ and
  broadcasts it.
  \State Each $P_i$ samples $\vb_i$ and sends $C_i = C_a * \vb_i +
  \Enc(0)$ to $P_1$.
  \label{step:mult}
  \State $P_1$ computes $C_c = \sum_i C_i$ and broadcasts it.
  \State The parties run distributed the decryption to obtain the
  share $\vc_i$.
\end{algorithmic}
\end{algorithm}

By definition,
\begin{align*}
  \sum \vc_i &= \Dec(C_c) \\
             &= \Dec(\sum_i C_a * \vb_i + \Enc(0)) \\
             &= \Dec(C_a * \sum_i \vb_i) \\
             &= \Dec(\sum_i \Enc(\va_i) * \sum_i \vb_i) \\
             &= (\sum_i \va_i) * (\sum_i \vb_i).
\end{align*}
This proves correctness. Furthermore, the distributed key setup hides
encrypted values from all parties, and the addition of $\Enc(0)$ in
step \ref{step:mult} ensures that $P_1$ only receives a fresh
encryption of $\va * \vb_i$.

\subsection{Replacing Homomorphic Encryption by a Dealer}
\label{sec:dealer}

It is possible to generate all correlated randomness (triples, random bits,
(e)daBits etc.) by one party. The security then relies on the fact that either
the dealer is trusted or all of the remaining parties. The offline phase
simply consists of the dealer generating the necessary randomness (for
example, a random triple $(a, b, ab)$) and then sending an additive secret
sharing to the remaining parties.

 \section{High-Level Secure Computation Building Blocks}
\label{sec:blocks}

In this section, we will discuss how to implement computation with MPC
with a focus on how it differs from computation on CPUs or GPUs.  Most
of the techniques below are already known individually. To the best of
our knowledge, however, we are the first to put them together in an
efficient and extensible framework for secure training of deep neural networks.

\paragraph{Oblivious Selection}

Plain secure computation does not allow branching because the parties
would need to be aware of which branch is followed. Conditional assignment
can be implemented as follows. If $b \in \{0,1\}$ denotes the
condition, $x + b \cdot (y - x)$ is either $x$ or $y$ depending on
$b$. If the condition is available in binary secret sharing but $x$
and $y$ in arithmetic secret sharing, $b$ has to be converted to the
latter. This can be done using a daBit
\citep{INDOCRYPT:RotWoo19}, which is a secret random bit shared both
in arithmetic and binary. It allows to mask a bit in one world by
XORing it. The result is then revealed and the masking is undone in
the other world.

\paragraph{Division}

\citet{FC:CatSax10} show how to implement quantized division
using the algorithm by \citet{Goldschmidt}. It mainly uses arithmetic
and the probabilistic truncation already explained. In addition, the
initial approximation requires a full bit decomposition as described above.
The error of the output depends on the error in the multiplications
used for Goldschmidt's iteration, which compounds in particular when
using probabilistic truncation. Due to the nature of secure
computation, the result of division by zero is undefined. One could
obtain a secret failure bit by testing the divisor to zero. However,
that is unnecessary in our algorithm, because so far we
only use division by secret value for the softmax function, where
the divisor is guaranteed to strictly positive.

\paragraph{Logarithm}

Computation logarithm with any public base can be reduced to logarithm
to base two using $\log_x y = \log_2 y \cdot \log_x 2$.
\citet{ACNS:AlySma19} propose to represent $y$ as $y = a \cdot 2^b$,
where $a \in [0.5,1)$ and $b \in \mathbb{Z}$. This then allows to
compute $\log_2 y = \log_2 a + b$. Given the restricted range of
$a$, $\log_2 a$ can be approximated using a division of polynomials.
Numerical stability and input range control are less of an issue here,
because we only use logarithm for the loss computation, which does not
influence the training.

\paragraph{Inverse square root}
\label{sec:isqrt}

\begin{figure}[tp!]
\begin{minipage}{\columnwidth}
\begin{algorithm}[H]
\caption{Separation ($\mathsf{Sep}$) \citep{inv-sqrt}}
\label{alg:sep}
\begin{algorithmic}[1]
  \Input{Secret share $\fpshare{\tilde{x}}$ where
    $2^{-f+1} \le \tilde{x} \le 2^{f-1}$.}
  \Output{\fpshare{\tilde{u}} and
    $\binshare{z_0}, \dots, \binshare{z_{k-1}}$ such that
    $z_{e+f} = 1$, $z_i = 0$ for $i \ne e+f$, and
    $\tilde{u} = \tilde{x}^{-1} \cdot 2^{e+1} \in [0.25,0.5)$.}

  \State $\binshare{z_0}, \dots, \binshare{z_{k-1}} \asn \mathsf{NP2}(\fpshare{\tilde{x}})$
  \State $\fpshare{2^{e-1}} \asn \mathsf{B2A}(\binshare{z_{2f-1}}, \dots, \binshare{z_0})$
  \State $\fpshare{\tilde{u}} \asn \fpshare{\tilde{x}} \cdot \fpshare{2^{e-1}}$
  \State \Return $\fpshare{\tilde{u}}, (\binshare{z_0}, \dots, \binshare{z_{k-1}}$
\end{algorithmic}
\end{algorithm}
\end{minipage}
\begin{minipage}{\columnwidth}
\begin{algorithm}[H]
\caption{Square-root compensation ($\mathsf{SqrtComp}$)}
\label{alg:sqrtcomp}
\begin{algorithmic}[1]
  \Input{Secret share $\binshare{z_0}, \dots, \binshare{z_{k-1}}$ such
    that $z_{e+f} = 1$, $z_i = 0$ for $i \ne e+f$}
  \Output{\fpshare{2^{-(e-1)/2}}}

  \State $k' \asn k/2$, $f' \asn f/2$, $c_0 \asn 2^{f/2+1}$, $c_1 = 2^{(f+1)/2+1}$
  \For{i = 0, \dots, k'}
  \State $\binshare{a_i} \asn \binshare{z_{2i}} \vee \binshare{z_{2i+1}}$
  \Comment{only $a_{e'} = 1$ for $e' = \floor{(e+f)/2}$}
  \EndFor
  \State $\fpshare{2^{-e'-2)}} \asn \mathsf{B2A}(\binshare{a_{2f'-1}}, \dots, \binshare{a_0})$
  \State{$\binshare{b} = \bigoplus_{i=0}^{k'-1} \binshare{z_{2i}}$}
  \Comment{$b = \mathsf{LSB}(e + f)$}
  \label{step:lsb}
  \State \Return $\fpshare{2^{-(e-1)/2}} \asn \mathsf{MUX}(c_0, c_1, \binshare{b}) \cdot \fpshare{2^{-e'-2}}$
\end{algorithmic}
\end{algorithm}
\end{minipage}
\begin{minipage}{\columnwidth}
\begin{algorithm}[H]
\caption{$\mathsf{InvertSqrt}$ \citep{inv-sqrt}}
\label{alg:invert-sqrt}
\begin{algorithmic}[1]
  \Input{Share \fpshare{\tilde{x}} where $2^{-f+1} \le \tilde{x} \le
    2^{f-1}$ where $2^{e-1} \le \tilde{x} \le 2^e$ for some $e \in
    \mathbb{Z}$}
  \Output{Share \fpshare{\tilde{y}} such $\tilde{y} \approx
    1/\sqrt{\tilde{x}}$}
  \State $\fpshare{\tilde{u}}, \binshare{z_0}, \dots,
  \binshare{z_{k-1}} \asn \mathsf{Sep}(\fpshare{\tilde{x}})$
  \Comment{$z_{e+f} = 1$}
  \State $\fpshare{\tilde{c}} \asn 3.14736 +
  \fpshare{\tilde{u}} \cdot (4.63887 \cdot
  \fpshare{\tilde{u}} - 5.77789)$
  \State $\fpshare{2^{-(e-1)/2}} \asn \mathsf{SqrtComp}(\binshare{z_0}, \dots,
  \binshare{z_{k-1}})$
  \State \Return $\fpshare{\tilde{c}} \cdot
  \fpshare{2^{-(e-1)/2}}$
\end{algorithmic}
\end{algorithm}
\end{minipage}

\begin{minipage}{\columnwidth}
\begin{algorithm}[H]
  \caption{Procedures in Algorithms \ref{alg:sep}--\ref{alg:invert-sqrt}}
\begin{description}
\item[$\binshare{x_0}, \dots, \binshare{x_{k-1}} \asn
  \mathsf{NP2}(\fpshare{x})$] Next power of two. This returns a
  one-hot vector
  of bits such indicating the closes larger power of two. We adapt
  Protocol 3.5 by \citet{FC:CatSax10} for this.
\item[$\intshare{x} \asn \mathsf{B2A}(\binshare{x_0}, \dots,
  \binshare{x_{k-1}})$] Domain conversion from binary to arithmetic as
  above.
\item[$\fpshare{c_b} \asn \mathsf{MUX}(c_0, c_1, \binshare{b})$]
  Oblivious selection as in \appref{sec:blocks}.
\end{description}
\end{algorithm}
\end{minipage}
\end{figure}

\citet{ACNS:AlySma19} propose to compute square root using
Goldschmidt and Raphson-Newton iterations. We could combine this with
the division operator introduced above. However, \citet{inv-sqrt} propose a more
direct computation that avoids running two successive iterations.
We have optimized their algorithm $\mathsf{SqrtComp}$ computing the
square root of a power of two as shown in Algorithm~
\ref{alg:sqrtcomp}.
As an essential optimization, we compute the least
significant bit of $e+f$ by simply XORing every other $z_i$ in step
\ref{step:lsb} of Algorithm~\ref{alg:sqrtcomp}. Given
that XOR does not require communication with binary secret sharing,
this is much more efficient than computing the least significant bit
of a sum of oblivious selections done by \citeauthor{inv-sqrt}
Remarkably, our optimization cuts the cost by roughly half. This is
due to the reduction in conversions from binary to arithmetic.
Furthermore, we correct some issues which result in outputs
that are off by a multiplication with a power of two.
We present the rest of their approach in Algorithms \ref{alg:sep} and
\ref{alg:invert-sqrt} (unchanged) for completeness.
Unlike \citet{inv-sqrt}, we do not explicitly state the truncation
after fixed-point multiplication.

\begin{table}
  \centering
  \tcaption{Total communication in kbit for inverse square root across a range
    of security models with one corrupted party. ``SH'' stands for semi-honest
    security and ``Mal.'' for malicious security.}
  \label{table:isqrt}
  \begin{tabular}{lrrrr}
    \toprule
    & \multicolumn{2}{c}{3 Parties} & \multicolumn{2}{c}{2 Parties} \\
    & SH & Mal. & SH & Mal. \\
    \midrule
    \citet{inv-sqrt}  & 19 & 160 & 481 & 25,456 \\
    Ours  & 9 & 114 & 342 & 21,522 \\
    \bottomrule
  \end{tabular}
\end{table}

Table \ref{table:isqrt} shows how our approach compares to \citet{inv-sqrt}.
In the context of the Figures in Table \ref{table:sec-models}, this implies an
improvement of up one quarter. The importance of inverse square root stems
from the fact that it is computed for every iteration and trainable parameter
in the parameter update.

\paragraph{Uniformly random fractional number}

Limiting ourselves to intervals of the form $[x, x + 2^e)$ for a
potentially negative integer $e$, we can reduce the problem to generate a
random $(f + e)$-bit number where $f$ is the fixed-point
precision. Recall that we represent a fractional number $x$ as
$\round{x \cdot 2^{-f}}$. Generating a random $n$-bit number is
straightforward using random bits, which in our protocol can be
generated as presented by \citet{SP:DEFKSV19}. However,
\citet{USENIX:DalEscKel21} and \citet{C:EGKRS20} present more
efficient approaches that involve mixed-circuit computation.

\paragraph{Communication cost}

Table \ref{table:cost} shows the total communication cost of some of
the building blocks in our three-party protocol for $f=16$. This setting
mandates the modulus $2^{64}$ because the division protocol requires
a bit length of $4f$.

\begin{table}[H]
  \centering
  \tcaption{Communication cost of select computation for $f=16$ and integer modulus $2^{64}$.}
  \label{table:cost}
  \begin{tabular}{lr}
    \toprule
    & Bits \\
    \midrule
    Integer multiplication & 192 \\
    Probabilistic truncation & 960 \\
    Nearest truncation & 2,225 \\
    Comparison & 668 \\
    Division (prob. truncation) & 10,416 \\
    Division (nearest truncation) & 24,081 \\
    Exponentiation (prob. truncation) & 16,303 \\
    Exponentiation (nearest truncation) & 43,634 \\
    Invert square root (prob. truncation) & 9,455 \\
    Invert square root (nearest truncation) & 15,539 \\
    \bottomrule
 \end{tabular}
\end{table}

\section{Deep Learning Building Blocks}\label{sec:mlblocks}

In this section, we will use the building blocks in Appendix~\ref{sec:blocks} to
construct high-level computational modules for deep learning.

\paragraph{Fully connected layers}

Both forward and back-propagation of fully connected layers can be
seen as matrix multiplications and thus can be implemented using dot
products. A particular challenge in secure computation is to compute a
number of outputs in parallel in order to save communication rounds.
We overcome this challenge by having a dedicated infrastructure in our
implementation that computes all dot products for a matrix
multiplication in a single batch of communication, thus reducing the
number of communication rounds.

\paragraph{2D convolution layers}

Similar to fully connected layers, 2D convolution and its corresponding
gradient can be implemented
using only dot products, and we again compute several output values
in parallel.

\paragraph{ReLU}

Given the input $x\in\Re$, a rectified linear unit (ReLU~\citealt{relu}) outputs
\[ \rl(x) \defeq
\max(x,0) = \iverson{x>0} \cdot x.
\]

It can thus be implemented as a comparison followed by an oblivious
selection. For back-propagation, it is advantageous to reuse the
comparison results from forward propagation due to the relatively high
cost of secure computation. Note that the comparison results are stored in
secret-shared form and thus there is no reduction in security.

\paragraph{Max pooling}

Similar to ReLU, max pooling can be reduced to comparison and
oblivious selection. In secure computation, it saves communication
rounds if the process uses a balanced tree rather than iterating over
all input values of one maximum computation. For example,
two-dimensional max pooling with a 2x2 window requires computing the
maximum of four values. We compute the maximum of two pairs of values
followed by the maximum of the two results.
For back-propagation, it
again pays off to store the intermediate results from forward
propagation in secret-shared form.

\paragraph{Dropout}

Dropout layers~\citep{dropout} require generating a random bit according to some
probability and oblivious selection. For simplicity, we only support
probabilities that are a power of two.

\paragraph{Batch normalization}

As another widely used deep learning module, batch normalization (BN, \citealt{bn}).
Based on a mini-batch $\mathcal{B}=\{x_i\}$ of any input statistic $x$,
the BN layer outputs
\begin{align}
    y_i
    = \gamma \frac{x_i-\mu_{\mathcal{B}}}{\sqrt{\sigma_{\mathcal{B}}^2}+\epsilon} + \beta,
\end{align}
where $\mu_{\mathcal{B}}$ and $\sigma^2_{\mathcal{B}}$ are the first and second
central moments of $x$, and $\gamma$, $\beta$ and $\epsilon$ are hyper-parameters.
In addition to basic arithmetic, it requires inverse square root, which is not
trivial in MPC.

\paragraph{Softmax output and cross-entropy loss}

For classification tasks,
consider softmax output units $\frac{\exp({x}_i)}{\sum_j \exp(x_j)}$,
where $\bm{x}=(x_1,x_2,\cdots)$ is the linear output (logits) of the last layer,
and the cross-entropy loss
$\ell= - \sum_{i} y_i \log \frac{\exp({x}_i)}{\sum_j \exp(x_j)}
= - \sum_{i} x_iy_i + \log\sum_{j} \exp( x_j )$,
where $\bm{y}=(y_1,y_2,\cdots)$ is the ground truth one-hot vector.
This usual combination requires computing the following gradient for
back-propagation:
\begin{align}\label{eq:grad}
\bigtriangledown_i \defeq
\frac{\partial\ell}{\partial x_i}&= \frac{\partial}{\partial x_i} \Big(
                                      -\sum_k y_k \cdot x_k +
                                      \log \sum_j \exp( x_j ) \Big)\nonumber\\
                                 &= -y_i + \frac{\exp(x_i)}{\sum_j \exp(x_j)}.
\end{align}

On the RHS of the above eq.~(\ref{eq:grad}),
the values in the denominator are potentially large due to the
exponentiation. This is prone to numerical overflow in our quantized representation,
because the latter puts strict limits on the values. We therefore
optimize the computation by first taking the maximum of the input values:
$x_{\max} = \max_i x_i$. Then we evaluate $\bigtriangledown_i$ based on
\begin{equation*}
\bigtriangledown_i
=
\frac{\exp(x_i-x_{\max})}{\sum_j \exp(x_j-x_{\max})} - y_i.
\end{equation*}
As $\forall{j}$, $x_j\le{}x_{\max}$,
we have $1 \le \sum_j \exp(x_j-x_{\max})\le L$,
where $L$ is the number of class labels ($L=10$ for MNIST).
Hence numerical overflow is avoided.
The same technique can be used to compute the sigmoid activation function, as
$\mathrm{sigmoid}(x) \defeq \frac{1}{1+\exp(-x)} = \frac{\exp(0)}{\exp(0)+\exp(-x)}$
is a special case of softmax.

\paragraph{Stochastic gradient descent}

The model parameter update in SGD only involves basic arithmetic:
\begin{equation}\label{eq:sgd}
\theta_j
\leftarrow
\theta_j - \frac{\gamma}{B} \sum_{i=1}^B \frac{\partial\ell_i}{\partial\theta_j},
\end{equation}
where $\theta_j$ is the parameter indexed by $j$,
 $\gamma>0$ is the learning rate, $B$ is the mini-batch size,
$\ell_i$ is the cross-entropy loss with respect to the $i$'th sample in the mini-batch.
To tackle the limited precision with quantization, we \emph{defer dividing by
the batch size $B$} to the model update in eq.~(\ref{eq:sgd}).
In other words, the back-propagation computes the gradient
$\frac{\partial\ell_i}{\partial\theta_j}$,
where the back-propagated error terms are not divided by $B$.
Division by the batch size $B$ only happens when the parameter update is performed.
Since the batch size is a power of two (128), it
is sufficient to use probabilistic truncation instead of full-blown
division. This both saves time and decreases the error.

\paragraph{Adam~\citep{adam} and AMSGrad~\citep{amsgrad}}

Optimizers in this category perform a more sophisticated parameter
update rule\footnote{Our implementation is slightly different from the
original Adam and AMSGrad, as we put $\epsilon$ inside the square root.
This is because the inverse square root is implemented as a basic operation
that can be efficiently computed in MPC.}:
\[
\theta_j \leftarrow \theta_j - \frac{\gamma}{\sqrt{v_j + \epsilon}} g_j,
\]
where
$\epsilon>0$ is a small constant to prevent division by zero,
and $g_j$ and $v_j$ are the first and second moments
of the gradient
$\frac{\partial\ell}{\partial\theta_j}
=\frac{1}{B}\sum_{i=1}^B
\frac{\partial\ell_i}{\partial\theta_j}$, respectively.
Division by the batch size $B$ can be skipped
in computing $\frac{\partial\ell}{\partial\theta_j}$,
because scaling of the gradient leads to the same scaling factor of $g_j$ and $\sqrt{v_j}$.
Both $g_j$ and $v_j$ are computed from the back-propagation result
using simple arithmetic and comparison (in the case of AMSGrad).
We compute the inverse square root as described in
Section~\ref{sec:blocks} above.

\paragraph{Parameter initialization}

\newcommand{\din}{d_{\mathrm{in}}}
\newcommand{\dout}{d_{\mathrm{out}}}
We use the widely adopted initialization method proposed by \citet{glorot2010understanding}.
Each weight $w$ between two layers with size $\din$ and $\dout$ is initialized by
\begin{displaymath}
w\sim \uniform\left[-\sqrt{\frac{6}{\din+\dout}}, \sqrt{\frac{6}{\din+\dout}}\right],
\end{displaymath}
where $\uniform[a, b]$ means the uniform distribution in the given range $[a,b]$.
Besides basic operations, it involves generating a uniformly distributed
random fractional value in a given interval, which is introduced in the previous section~\ref{sec:blocks}.
All bias values are initialized to 0.

 \section{Models}
\label{sec:model-code}

The neural network structures investigated in this paper are given by the following
Figures~\ref{fig:net-a}, \ref{fig:net-b}, \ref{fig:net-c}, \ref{fig:net-d} and \ref{fig:alex}.
The structure of each network is formatted in Keras\footnote{https://keras.io} code.

\begin{figure*}[ht!]
  \begin{verbatim}
    tf.keras.layers.Flatten(),
    tf.keras.layers.Dense(128, activation='relu'),
    tf.keras.layers.Dense(128, activation='relu'),
    tf.keras.layers.Dense(10, activation='softmax')
\end{verbatim}
  \caption{Network A used by \citet{SP:MohZha17}}
  \label{fig:net-a}
\end{figure*}

\begin{figure*}[ht!]
  \begin{verbatim}
    tf.keras.layers.Conv2D(16, 5, 1, 'same', activation='relu'),
    tf.keras.layers.MaxPooling2D(2),
    tf.keras.layers.Conv2D(16, 5, 1, 'same', activation='relu'),
    tf.keras.layers.MaxPooling2D(2),
    tf.keras.layers.Flatten(),
    tf.keras.layers.Dense(100, activation='relu'),
    tf.keras.layers.Dense(10, activation='softmax')
\end{verbatim}
  \caption{Network B used by \citet{CCS:LJLA17}}
  \label{fig:net-b}
\end{figure*}

\begin{figure*}[ht!]
  \begin{verbatim}
    tf.keras.layers.Conv2D(20, 5, 1, 'valid', activation='relu'),
    tf.keras.layers.MaxPooling2D(2),
    tf.keras.layers.Conv2D(50, 5, 1, 'valid', activation='relu'),
    tf.keras.layers.MaxPooling2D(2),
    tf.keras.layers.Flatten(),
    tf.keras.layers.Dropout(0.5),
    tf.keras.layers.Dense(100, activation='relu'),
    tf.keras.layers.Dense(10, activation='softmax')
\end{verbatim}
  \caption{Network C used by \citet{lecun1998gradient} (with optional Dropout layer)}
  \label{fig:net-c}
\end{figure*}

\begin{figure*}[ht!]
  \begin{verbatim}
    tf.keras.layers.Conv2D(5, 5, 2, 'same', activation='relu'),
    tf.keras.layers.Flatten(),
    tf.keras.layers.Dense(100, activation='relu'),
    tf.keras.layers.Dense(10, activation='softmax')
\end{verbatim}
  \caption{Network D used by \citet{ASIACCS:RWTSSK18}}
  \label{fig:net-d}
\end{figure*}

\begin{figure*}[ht!]
  \begin{verbatim}
  #1st Convolutional Layer
  Conv2D(filters=96, input_shape=(32,32,3), kernel_size=(11,11), strides=(4,4),
    padding=9),
  Activation('relu'),
  MaxPooling2D(pool_size=3, strides=(2,2)),
  BatchNormalization(),

  #2nd Convolutional Layer
  Conv2D(filters=256, kernel_size=(5, 5), strides=(1,1), padding=1),
  Activation('relu'),
  BatchNormalization(),
  MaxPooling2D(pool_size=(2,2), strides=1),

  #3rd Convolutional Layer
  Conv2D(filters=384, kernel_size=(3,3), strides=(1,1), padding=1),
  Activation('relu'),

  #4th Convolutional Layer
  Conv2D(filters=384, kernel_size=(3,3), strides=(1,1), padding=1),
  Activation('relu'),

  #5th Convolutional Layer
  Conv2D(filters=256, kernel_size=(3,3), strides=(1,1), padding=1),
  Activation('relu'),

  #Passing it to a Fully Connected layer
  # 1st Fully Connected Layer
  Dense(256),
  Activation('relu'),

  #2nd Fully Connected Layer
  Dense(256),
  Activation('relu'),
\end{verbatim}
\end{figure*}
\begin{figure*}[th!] \begin{verbatim}
  #Output Layer
  Dense(10)
  \end{verbatim}
  \caption{AlexNet for CIFAR-10 by \citet{PoPETS:WTBKMR21}. Note that our implementation allows any padding unlike Keras.}
  \label{fig:alex}
\end{figure*}

 \section{Fashion MNIST}

\begin{figure*}[ht]
  \centering
  \pgfplotsset{
    every axis legend/.append style={
      at={(1.2,1)},
      anchor=north west,
    },
  }
  \begin{tikzpicture}
    \begin{axis}[
ylabel = training loss,
      xlabel = \#epochs,
ytick pos = left,
legend style = {cells={anchor=west}},
      legend entries = {
        {secure, SGD, $\gamma=0.01$},
        {secure, AMSGrad, $\gamma=0.001$},
        {cleartext, SGD, $\gamma=0.01$},
        {cleartext, AMSGrad, $\gamma=0.001$},
      }
      ]
      \addplot[colour1] table[x=epoch, y=loss, mark=none]{data/trunc_pr-fashion};
      \addplot[colour3,densely dotted] table[x=epoch, y=loss, mark=none]{data/amsgrad-rate.001-trunc_pr-fashion};
      \addplot[colour5,densely dashdotted] table[x index=0, y index=2, mark=none]{data/log-C-50-.01-fashion};
      \addplot[colour4,densely dashdotdotted] table[x index=0, y index=2, mark=none]{data/log-C-50-amsgrad-fashion};
    \end{axis}
    \begin{axis}[
      ylabel = testing error,
      ylabel near ticks,
      ytick pos = right,
      y dir = reverse,
yticklabel style={/pgf/number format/.cd,fixed,precision=2},
      ]
      \addplot[colour1] table[x=epoch, y=error, mark=none]{data/trunc_pr-fashion};
      \addplot[colour3,densely dotted] table[x=epoch, y=error, mark=none]{data/amsgrad-rate.001-trunc_pr-fashion};
      \addplot[colour5,densely dashdotted] table[x index=0, y expr=1-\thisrowno{8}, mark=none]{data/log-C-50-.01-fashion};
      \addplot[colour4,densely dashdotdotted] table[x index=0, y expr=1-\thisrowno{8}, mark=none]{data/log-C-50-amsgrad-fashion};
    \end{axis}
  \end{tikzpicture}
  \caption{Learning curves for image classification using network C on the Fashion MNIST dataset, with $f=16$ and probabilistic truncation. $\gamma$ is the learning rate.}
  \label{fig:fashion}
\end{figure*}
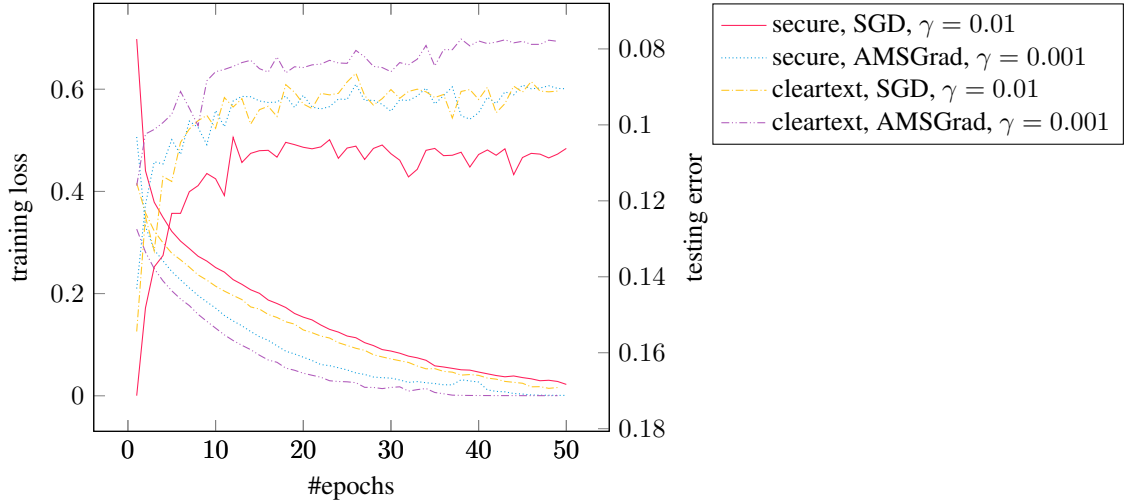

We run our implementation on Fashion MNIST~\citep{fashion} for a more complete
picture. Figure~\ref{fig:fashion} shows our results. See
\citet{fashion} for an overview on how other models perform in
cleartext.

 \section{More Experimental Results}
\label{sec:more-figures}

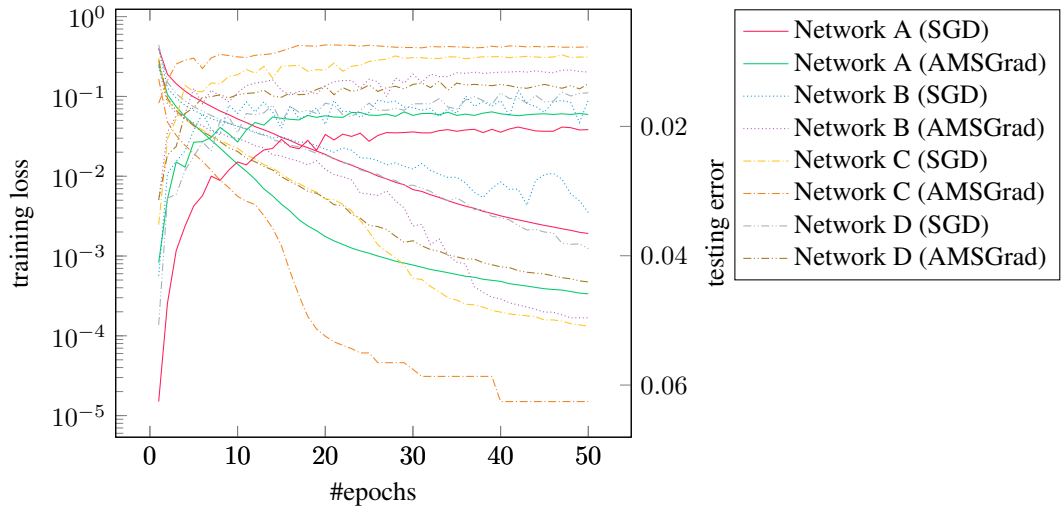
\begin{figure}[htbp]
  \centering
  \pgfplotsset{
every axis legend/.append style={
  at={(1.2,1)},
  anchor=north west,
},
}
{
\begin{tikzpicture}
\begin{axis}[
ylabel = training loss,
  xlabel = \#epochs,
ymode = log,
  ytick pos = left,
legend style = {cells={anchor=west}},
  legend entries = {
    {Network A (SGD)},
    {Network A (AMSGrad)},
    {Network B (SGD)},
    {Network B (AMSGrad)},
    {Network C (SGD)},
    {Network C (AMSGrad)},
    {Network D (SGD)},
    {Network D (AMSGrad)},
  },
  ]
  \addplot[colour1] table[x=epoch, y=loss, mark=none]{data/A-trunc_pr};
  \addplot[colour2] table[x=epoch, y=loss, mark=none]{data/A-amsgrad-rate.1-trunc_pr};
  \addplot[colour3,densely dotted] table[x=epoch, y=loss, mark=none]{data/B-trunc_pr};

  \addplot[colour4,densely dotted] table[x=epoch, y=loss, mark=none]{data/B-amsgrad-rate.1-trunc_pr};
  \addplot[colour5,densely dashdotted] table[x=epoch, y=loss, mark=none]{data/trunc_pr};
  \addplot[colour6,densely dashdotted] table[x=epoch, y=loss, mark=none]{data/amsgrad-rate.001-trunc_pr};
  \addplot[colour7,densely dashdotdotted] table[x=epoch, y=loss, mark=none]{data/D-2dense-trunc_pr};
  \addplot[colour8,densely dashdotdotted] table[x=epoch, y=loss, mark=none]{data/D-2dense-amsgrad-rate.1-trunc_pr};
\end{axis}
\begin{axis}[
  ylabel = testing error,
  ylabel near ticks,
  ytick pos = right,
  y dir = reverse,
yticklabel style={/pgf/number format/.cd,fixed,precision=2},
  scaled y ticks = false,
  ]
  \addplot[colour1] table[x=epoch, y=error, mark=none]{data/A-trunc_pr};
  \addplot[colour2] table[x=epoch, y=error, mark=none]{data/A-amsgrad-rate.1-trunc_pr};
  \addplot[colour3,densely dotted] table[x=epoch, y=error, mark=none]{data/B-trunc_pr};

  \addplot[colour4,densely dotted] table[x=epoch, y=error, mark=none]{data/B-amsgrad-rate.1-trunc_pr};
  \addplot[colour5,densely dashdotted] table[x=epoch, y=error, mark=none]{data/trunc_pr};
  \addplot[colour6,densely dashdotted] table[x=epoch, y=error, mark=none]{data/amsgrad-rate.001-trunc_pr};
  \addplot[colour7,densely dashdotdotted] table[x=epoch, y=error, mark=none]{data/D-2dense-trunc_pr};
  \addplot[colour8,densely dashdotdotted] table[x=epoch, y=error, mark=none]{data/D-2dense-amsgrad-rate.1-trunc_pr};
\end{axis}
\end{tikzpicture}
}
\caption{Loss and accuracy for various networks, $f=16$, and probabilistic truncation.\label{fig:nets}}

 \end{figure}

For the task of classifying MNIST digits, Figure~\ref{fig:nets} shows the
learning curves for all networks used by \citet{PoPETS:WagGupCha19}.
Among all the investigated structures, LeNet (network C) performs the best.
Furthermore, AMSGrad consistently improves the classification performance as compared to SGD.

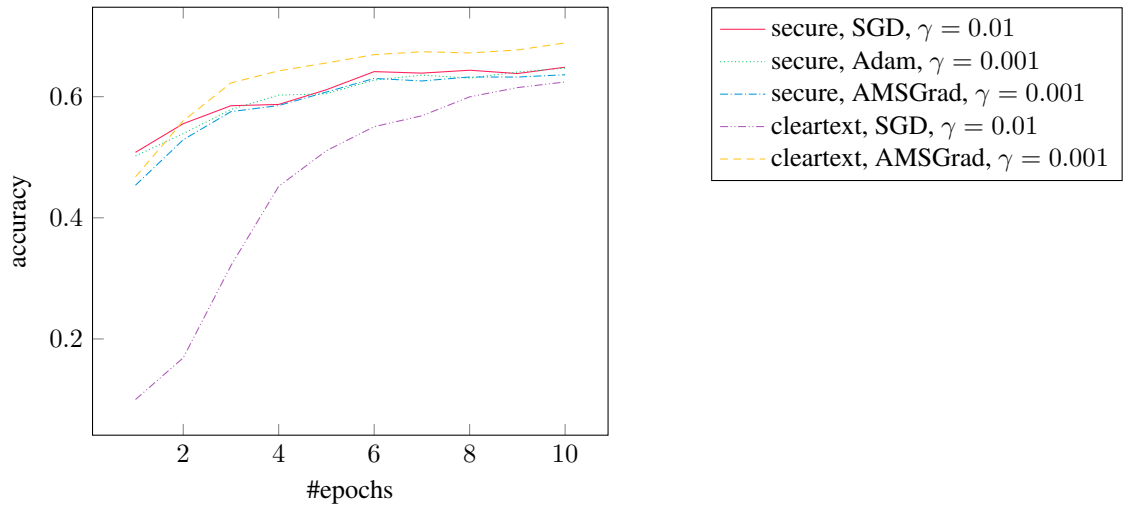
\begin{figure}
\centering
\pgfplotsset{
every axis legend/.append style={
  at={(1.2,1)},
  anchor=north west,
},
}
{
\begin{tikzpicture}
\begin{axis}[
ylabel = accuracy,
  xlabel = \#epochs,
ytick pos = left,
legend style = {cells={anchor=west}},
  legend entries = {
    {secure, SGD, $\gamma=0.01$},
    {secure, Adam, $\gamma=0.001$},
    {secure, AMSGrad, $\gamma=0.001$},
    {cleartext, SGD, $\gamma=0.01$},
    {cleartext, AMSGrad, $\gamma=0.001$},
  },
  ]
  \addplot[colour1] table[header=false, x expr=1+\coordindex, y index=0, mark=none]{data/cifar-secret-sgd};
  \addplot[colour2,densely dotted] table[header=false, x expr=1+\coordindex, y index=0, mark=none]{data/cifar-secret-adam};
  \addplot[colour3,densely dashdotted] table[header=false, x expr=1+\coordindex, y index=0, mark=none]{data/cifar-secret-amsgrad};
  \addplot[colour4,densely dashdotdotted] table[header=false, x expr=1+\coordindex, y expr=\thisrowno{0}/100, mark=none]{data/cifar-clear-sgd};
  \addplot[colour5,densely dashed] table[header=false, x expr=1+\coordindex, y expr=\thisrowno{0}/100, mark=none]{data/cifar-clear-adam};
\end{axis}
\end{tikzpicture}
}
\caption{Comparison of cleartext training and secure training for CIFAR-10 with $f=16$ and probabilistic truncation. $\gamma$ is the learning rate.\label{fig:comp-cifar10}}
\end{figure}

Figure~\ref{fig:comp-cifar10} shows the comparison of cleartext training and
secure training for CIFAR-10 with $f=16$ and probabilistic truncation.  Our
MPC implementation has gained similar performance with the cleartext counterpart.

  \section{Hyperparameter Settings}

In the following we discuss our choice of hyperparameters.

\begin{description}
\item[Number of epochs]  As we found convergence after  100 epochs, we
  have  run most  of our  benchmarks for  150 epochs,  except for  the
  comparison of optimizers where we stopped at 100.
\item[Early stop] We have not used early stop.
\item[Mini-batch size] We have used 128 throughout as it is a standard
  size. We briefly trialed 1024 as suggested by
  \citet{train-quantized}, but did not found any improvement.
\item[Reshuffling training samples] At the beginning of each epoch, we
  randomly re-shuffle the training samples using the Fisher-Yates shuffle
  with MP-SPDZ's internal pseudo-random number generator as randomness
  source.
\item[Learning rate] We have tried a number of learning rates as
  documented in Figure \ref{fig:c-optimizers}. As a result, we settled
  for 0.01 for SGD and 0.001 for AMSGrad in further benchmarks.
\item[Learning rate decay/schedule] We have not used either.
\item[Random initialization] The platform uses independent random
  initialization by design.
\item[Dropout] We have experimented with Dropout but not found any
  improvement as shown in Figure \ref{fig:c-optimizers}.
\item[Input preprocessing] We have normalized the inputs to $[0, 1]$.
\item[Test/training split] We have used the usual MNIST split.
\item[Hyperparameters for Adam/AMSGrad] We use the common choice
  $\beta_1 = 0.9$, $\beta_2 = 0.999$, and $\epsilon = 10^{-8}$.
\end{description}

 \section{List of Symbols}

\begin{table}[H]
\centering
\caption{List of Symbols}
\begin{tabular}{cl}
    \toprule
    $\Re$ & the set of real numbers\\
    $\Ze$ & the set of integers\\
    $x$, $y$, $\lambda$ & real numbers in $\Re$\\
    $\bm{x}$, $\bm{y}$ & real vectors \\
    $\round{x}$ & nearest rounding of $x$: $\round{x}\defeq\argmin_{z\in\Ze} \vert{}x-z\vert$\\
    $\floor{x}$ & floor function: $\floor{x}\defeq\max\{z\in\Ze\,\vert\,z\le{x}\} $\\
    $\{x\}$ & the fractional part of $x$: $\{x\}\defeq{}x-\floor{x}$\\
    $Q^f(x)$ & the integer representation of $x$: $Q^f(x)=\round{x\cdot2^f}$\\
    $\uniform(a,b)$ & uniform distribution on $[a,b]$\\
    $\bernoulli(p)$ & Bernoulli distribution (with probability $p$ the random variable equals $1$)\\
    $\iverson{c}$ & Iverson bracket ($\iverson{c}=1$ if $c$ is true otherwise 0)\\
    $\vert\cdot\vert$ & absolute value\\
    $\Vert\cdot\Vert$ & Frobenius norm\\
    $\mathrm{Prob}(\cdot)$ & probability of the parameter statement being true\\
    $(\mathrm{condition}~?~a~:~b)$ & oblivious selection\\
    \bottomrule
\end{tabular}
\end{table}

 \section{Proofs}\label{sec:proofs}

\subsection{Proof of Proposition \ref{thm:unbiased}}

\begin{proof}
    By Eq.~(\ref{eq:round}),
    \[
        E(b) = 0 \times (1-\{\mu\}) + 1 \times \{\mu\} = \{\mu\}.
    \]
    Therefore
    \begin{align*}
        E(R^f(xy)) &= E( \floor{\mu} + b )
                   = \floor{\mu} + E(b)
                   = \floor{\mu} + \{\mu\}
                   = \floor{\mu} + (\mu-\floor{\mu})
                   = \mu
                   = Q^f(x) Q^f(y) 2^{-f}.
    \end{align*}
    Notice that $Q^f(A)$ is simply obtained through entry-wise quantization $Q^f(a_{ij})$.
    By our matrix multiplication algorithm,
    \begin{align*}
        E(R_{ij}^f(AB)) &= E\left(\sum_{k}R^f(A_{ik}B_{kj})\right)\nonumber\\
                        &= \sum_{k} E( R^f(A_{ik}B_{kj}) ) \nonumber\\
                        &= \sum_{k} Q^f(A_{ik}) Q^f(B_{kj}) 2^{-f} \nonumber\\
                        &= 2^{-f} \sum_{k} Q_{ik}^f(A) Q_{kj}^f(B_{kj}) \nonumber\\
                        &= 2^{-f} ( Q^f(A) Q^f(B) )_{ij}.
    \end{align*}
    In matrix form, it gives
    \begin{equation*}
        E(R^f(AB)) = 2^{-f} Q^f(A) Q^f(B).
    \end{equation*}
\end{proof}

\subsection{A Lemma of Probabilistic Rounding}

\begin{lemma}\label{thm:bound}
Given $x,y\in(-2^k,2^k)\subset\Re$,
\begin{align*}
    \left\vert xy - R^f(xy) 2^{-f} \right\vert
    &
    <
    (\max\{\vert{}x\vert,\vert{}y\vert\} + 1)\epsilon + \epsilon^2/4
    <
    (2^k+1) \epsilon + \epsilon^2/4,
\end{align*}
where $\epsilon\defeq{}2^{-f}$.
\end{lemma}

\begin{proof}
    We have
    \[
        \vert xy - R^f(xy)2^{-f} \vert
        \le
        \vert xy - \mu 2^{-f} \vert + \vert R^f(xy)2^{-f} - \mu 2^{-f}\vert
        =
        \vert xy - \mu 2^{-f} \vert + 2^{-f} \vert R^f(xy)- \mu\vert.
    \]
    By the definition of $R^f(xy)$,
    \begin{align*}
        \vert R^f (xy)-\mu \vert
        = \vert \floor{\mu} + b - \mu\vert
        = \vert b - (\mu-\floor{\mu}) \vert
        = \vert b - \{\mu\} \vert
        < 1.
    \end{align*}
    Therefore, we have
    \[
        \vert xy - R^f(xy)2^{-f} \vert
        <
        \vert xy - \mu 2^{-f} \vert + 2^{-f}
        =
        \vert xy - \mu 2^{-f} \vert + \epsilon.
    \]
    Now we only need to bound the deterministic term $\vert xy - \mu 2^{-f} \vert$.
    We have
    \begin{align*}
        \vert xy - \mu 2^{-f} \vert
        &=
        \vert xy - Q^f(x) Q^f(y) 2^{-2f} \vert \nonumber\\
        &=
        \vert Q^f(x) Q^f(y) 2^{-2f} - xy \vert \nonumber\\
        &=
        \vert ( Q^f(x) 2^{-f} -x )( Q^f(y) 2^{-f} - y )
                + 2^{-f}Q^f(x)y + 2^{-f}Q^f(y)x - 2xy \vert \nonumber\\
        &=
        \vert ( Q^f(x) 2^{-f} -x )( Q^f(y) 2^{-f} - y )
                + (2^{-f}Q^f(x)-x)y + (2^{-f}Q^f(y)-y)x \vert \nonumber\\
        &\le
        \vert Q^f(x) 2^{-f} -x \vert \cdot \vert Q^f(y) 2^{-f} - y \vert
        +
        \vert 2^{-f}Q^f(x) - x \vert \cdot \vert{}y\vert
        +
        \vert 2^{-f}Q^f(y) - y \vert \cdot \vert{}x\vert.
    \end{align*}
    We notice that $\forall{x}\in\Re$,
    \begin{align*}
        \vert 2^{-f}Q^f(x) - x \vert
        &=
        \vert 2^{-f}\round{x\cdot2^f} - x \vert
        =
         2^{-f} \vert \round{x\cdot2^f} - x\cdot2^f \vert\nonumber\\
        &\le
         2^{-f} \times \frac{1}{2}
         = 2^{-f-1}
         = \frac{\epsilon}{2}.
    \end{align*}
    Then, we get
    \begin{align*}
        \vert xy - \mu 2^{-f} \vert
        &\le
        \frac{\epsilon}{2} \cdot \frac{\epsilon}{2}
        +
        \frac{\epsilon}{2} \cdot ( \vert x \vert + \vert y \vert )
        =
        \frac{\epsilon^2}{4}
        +
        \frac{\epsilon}{2} \cdot ( \vert x \vert + \vert y \vert )\nonumber\\
        &\le
        \frac{\epsilon^2}{4}
        +
        \frac{\epsilon}{2} \cdot ( 2 \max\{\vert{x}\vert, \vert{y}\vert\} )
        =
        \frac{\epsilon^2}{4}
        +
        \max\{\vert{x}\vert, \vert{y}\vert\} \epsilon.
    \end{align*}
    In summary,
    \begin{align*}
         \vert xy - R^f(xy)2^{-f} \vert
         <
         \vert xy - \mu 2^{-f} \vert + \epsilon
         \le
         \frac{\epsilon^2}{4} +
         \max\{\vert{x}\vert, \vert{y}\vert\} \epsilon
         +\epsilon
         =
         (\max\{\vert{x}\vert, \vert{y}\vert\}+1) \epsilon
         + \frac{\epsilon^2}{4}.
    \end{align*}
    The second ``$<$'' is trivial by noting our assumption
    \begin{align*}
    \max\{\vert{x}\vert, \vert{y}\vert\} < 2^k.
    \end{align*}
\end{proof}

\subsection{Proof of Proposition~\ref{thm:matbound}}

\begin{proof}
The proof is based on Lemma~\ref{thm:bound}.
    \begin{align*}
    \Vert
    AB - R^f(AB) 2^{-f}
    \Vert
    &=
    \sqrt{\sum_{i,j}
    \left( (AB)_{ij} - R^f_{ij}(AB) 2^{-f} \right)^2}
    \nonumber\\
    &=
    \sqrt{\sum_{i,j}
    \left( \sum_{l} A_{il} B_{lj}
           - \sum_{l} R^f(A_{il}B_{lj}) 2^{-f} \right)^2}
    \nonumber\\
    &=
    \sqrt{\sum_{i,j}
    \left\vert \sum_{l}
        \left(A_{il}B_{lj}
           - R^f(A_{il}B_{lj}) 2^{-f}
        \right)
    \right\vert^2}
    \nonumber\\
    &
    \le
    \sqrt{\sum_{i,j}
        \left(
            \sum_{l}
            \left\vert A_{il}B_{lj} - R^f(A_{il}B_{lj}) 2^{-f} \right\vert
        \right)^2}
    \nonumber\\
    &
    <
    \sqrt{\sum_{i,j}
        \left[
        \sum_{l} \left( (2^k+1)\epsilon + \frac{\epsilon^2}{4} \right)
        \right]^2 }
    \nonumber\\
    &
    =
    \sqrt{mp\cdot{}n^2\cdot\left( (2^k+1)\epsilon + \frac{\epsilon^2}{4} \right)^2 }
    \nonumber\\
    &
    =
    \sqrt{mp}n\left((2^k+1)\epsilon + \frac{\epsilon^2}{4}\right).
    \end{align*}
    Multiplying both sides by $2^f$, we get
    \begin{align*}
    \Vert R^f(AB) - 2^{f} A B \Vert
    <
    \sqrt{mp}n\left((2^k+1) + \frac{\epsilon}{4}\right).
    \end{align*}
\end{proof}

\subsection{Proof of Proposition~\ref{thm:std}}

\begin{proof}
    Note that the variance, denoted as $\sigma^2(\cdot)$, of $R^f(xy)$ is
    \begin{align*}
        \sigma^2( R^f(xy) )
        &=
        \sigma^2(b)
        =
        E(b^2) - (E(b))^2
        =
        E(b) - (E(b))^2
        =
        \{\mu\} - (\{\mu\})^2
        =
        \{\mu\} (1 - \{\mu\})
        \\
        &\le
        \left(\frac{ \{\mu\} + (1 - \{\mu\}) }{2}\right)^2
        =
        \frac{1}{4}.
    \end{align*}
    Therefore,
    \begin{align*}
        \sigma^2( R^f_{ij}(AB) )
        &=
        \sigma^2\left( \sum_k R^f(A_{ik} B_{kj}) \right)
        =
        \sum_k \sigma^2\left( R^f(A_{ik} B_{kj}) \right)\\
        &\le
        \sum_k \frac{1}{4}
        =
        \frac{n}{4}.
    \end{align*}
    In matrix form, the element-wise variance of the random matrix $R^f(AB)$ is
    \[
        \sigma^2(R^f(AB)) \le \frac{n}{4} \bm{1}\bm{1}^\T,
    \]
    where $\bm{1}$ is the column vector of 1's.
    Note all $mp$ entries in $R^f(AB)$ are independent. By the Chebyshev inequality,
    with probability at least $1-\frac{mp}{\rho}$, the following is true
        \[
        \frac{ \Vert R^f(AB) - E(R^f(AB)) \Vert^2 }{n/4}
        \le \rho,
        \]
    which can be re-written as
        \[
        \Vert R^f(AB) - E(R^f(AB)) \Vert \le \frac{\sqrt{\rho{n}}}{2}.
        \]
    Let $\iota\defeq\frac{1}{2}\sqrt{\frac{\rho}{mp}}$. Then $\rho=4\iota^2mp$.
    In conclusion, with probability at least
        \[
        1-\frac{mp}{\rho}
        =
        1-\frac{mp}{4\iota^2mp}
        =
        1-\frac{1}{4\iota^2},
        \]
    the following is true:
        \[
        \Vert R^f(AB) - E(R^f(AB)) \Vert \le \frac{\sqrt{\rho n}}{2}
        =
        \frac{\sqrt{4\iota^2mp\cdot{n}}}{2}
        =
        \iota\sqrt{mnp}.
        \]
        Based on Proposition~\ref{thm:unbiased}, the statement can be equivalently written as
        \[
        \Vert R^f(AB) - 2^{-f}Q^f(A)Q^f(B) \Vert \le \iota \sqrt{mnp}.
        \]
\end{proof}

\end{document}